\documentclass[10pt,twocolumn,letterpaper]{article}

\usepackage{cvpr}              % To produce the CAMERA-READY version
\usepackage{multirow}
\usepackage{tabularx}
\usepackage{booktabs}
\usepackage{algorithm}
\usepackage{algorithmic}

\definecolor{cvprblue}{rgb}{0.21,0.49,0.74}
\usepackage[pagebackref,breaklinks,colorlinks,allcolors=cvprblue]{hyperref}

\def\confName{CVPR}
\def\confYear{2027}

\title{Do Occupancy World Models Really Follow Actions?}

\author{Jongseong Bae\\
Yonsei University\\
Seoul, Republic of Korea\\
{\tt\small js.bae@yonsei.ac.kr}
\and
Ha Young Kim\thanks{Corresponding author.}\\
Yonsei University\\
Seoul, Republic of Korea\\
{\tt\small hayoung.kim@yonsei.ac.kr}
}

\begin{document}
\maketitle
\begin{abstract}
Occupancy world models predict future 3D scenes conditioned on ego actions,
yet their predictions may not faithfully follow the specified trajectory under
counterfactual actions.
Because driving datasets provide sparse coverage of alternative trajectories
for each scene, existing models learn the spatial effect of ego motion
implicitly from observed scene--trajectory combinations, which may
entangle action effects with scene--trajectory correlations.
We propose \textbf{ReCast}, an ego-trajectory-guided transformer that grounds
future prediction in action-induced geometric correspondence across
ego-centric frames.
ReCast decouples historical encoding from future decoding and aligns
deformable interactions using relative ego transformations, while retaining
flexibility for non-rigid scene evolution.
We also introduce a counterfactual action-faithfulness benchmark covering
in-distribution, tail, and out-of-distribution trajectories.
On Occ3D-nuScenes, ReCast achieves competitive standard
forecasting performance while improving counterfactual
forecasting, with particularly clear gains under
out-of-distribution trajectories.
\end{abstract}
    
\section{Introduction}
\label{1_intro}
World models~\cite{ha2018world} learn predictive representations of the surrounding environment, enabling embodied agents to anticipate how a scene may evolve in response to their actions, and have therefore been increasingly adopted in autonomous driving~\cite{yang2023unisim,hu2023gaia,wang2024driving}.
They have been explored using image-, LiDAR-, and occupancy-based representations.
Image-based approaches~\cite{hu2023gaia,wang2024drivedreamer,wang2024driving} benefit from the scalability of large-scale camera data but lack explicit 3D geometric structure, whereas LiDAR-based approaches~\cite{zhang2024copilot4d,zyrianov2025lidardm,wu2026gem} provide accurate spatial measurements but represent the environment through inherently sparse point observations.
\begin{figure}[hbt!]
    \centering
    \includegraphics[width=1\linewidth]{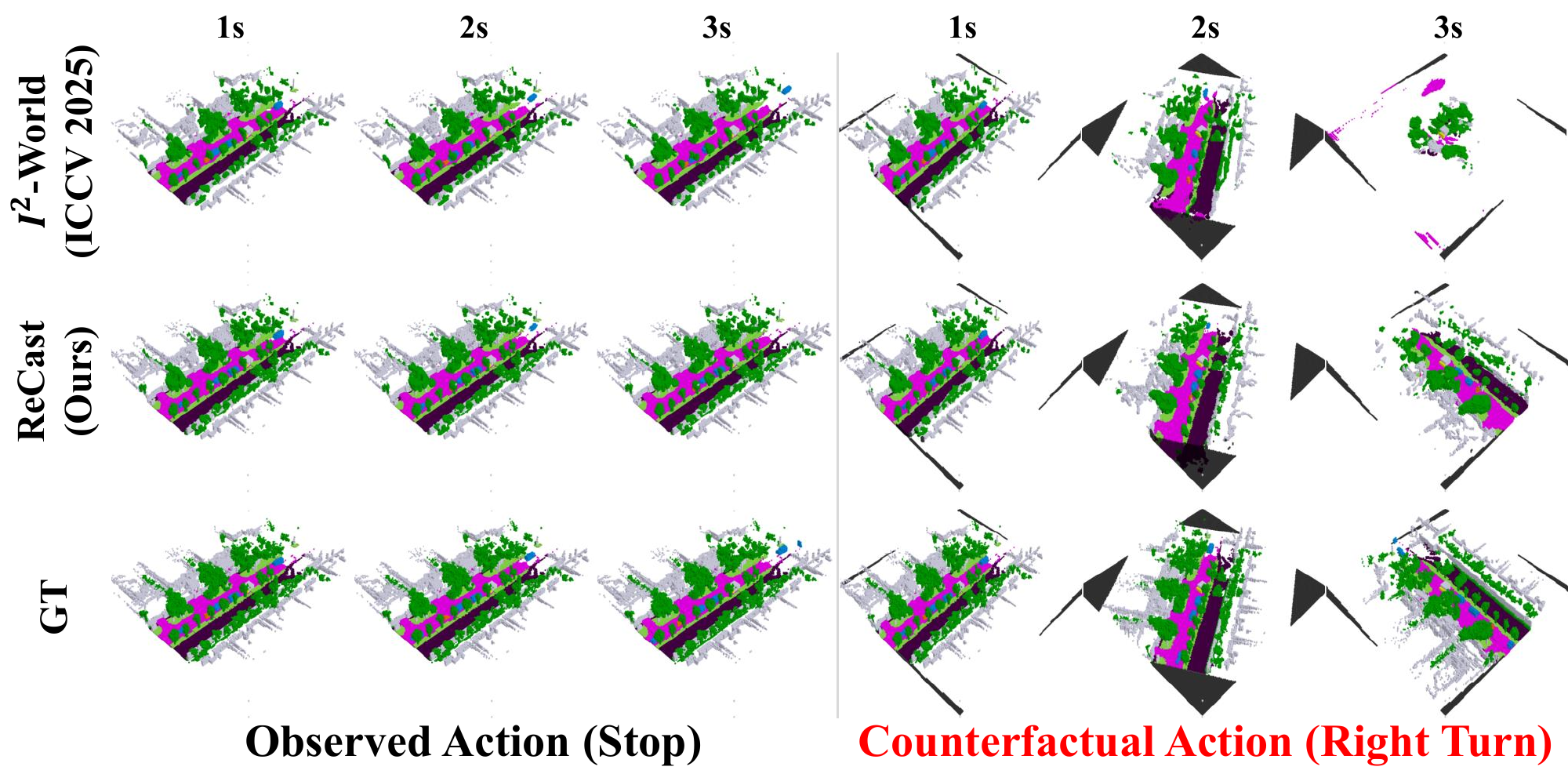}
\caption{\textbf{Action faithfulness under counterfactual ego trajectories.}
Both models produce plausible forecasts with the recorded trajectory.
Under an out-of-distribution counterfactual trajectory,
$I^2$-World does not fully adapt the scene layout to the geometric change
induced by ego motion.
In contrast, ReCast produces a forecast more consistently aligned with the
specified trajectory.}
    \label{fig1}
    % \vspace{-3.mm}
\end{figure}
3D occupancy has emerged as a promising alternative, as it jointly captures
physically grounded geometry and semantics within a unified volumetric space.

Recent occupancy-based world models~\cite{gu2024dome,liao2025i2}
formulate forecasting as action-conditioned scene generation from historical
observations.
They typically encode the ego trajectory as a learned condition and implicitly
infer the corresponding scene evolution from training data.
While promising, this formulation does not explicitly impose the geometric
correspondence induced by the specified action.
In driving datasets such as Occ3D-nuScenes~\cite{tian2023occ3d}, each scene is
paired with only its recorded trajectory, providing sparse coverage of
alternative scene--trajectory combinations.
Models can therefore learn action effects jointly with scene context, rather
than identifying how a specified trajectory geometrically transforms the scene
across contexts.
Here, a trajectory different from the recorded one for the same scene is a
counterfactual action, and action faithfulness indicates whether the predicted
scene geometrically follows the specified trajectory.
This scene-dependent reasoning is highly effective for nominal forecasting and can still produce plausible predictions for nearby counterfactual actions.
However, its geometric consistency may deteriorate when the specified
trajectory is paired with a scene context not jointly observed during
training, particularly as the action departs from the training distribution.
As shown in Fig.~\ref{fig1}, an out-of-distribution (OOD) counterfactual
trajectory can lead to a predicted scene layout that is geometrically
inconsistent with the specified ego motion.
% Models may therefore interpret the action jointly with scene context and partially rely on a scene-centric action prior, biasing their forecasts toward the trajectory likely for the observed scene. 
% As illustrated in Fig.~\ref{fig1}, this bias can emerge even when a scene is paired with an in-distribution but counterfactual action, weakening the geometric alignment between the specified action and the predicted scene layout. 
% This misalignment is further amplified as the specified trajectory moves toward the tail of, or beyond, the training-action distribution.

Unlike appearance-based representations, 3D occupancy is defined in an explicit metric coordinate system. Although an ego trajectory does not fully determine future scene evolution, it does determine the geometric correspondence of persistent scene content across ego-centric frames. We show that encoding this correspondence as an architectural inductive bias substantially improves action faithfulness, while leaving object motion, visibility changes, and newly revealed content to be inferred from temporal context. Rather than rigidly prescribing the future scene, action geometry structures how historical evidence is retrieved and integrated during full-scene forecasting. This perspective leads us to rethink occupancy world modeling from action-conditioned generation toward geometry-grounded, action-faithful forecasting.

In this work, we propose \textbf{ReCast}, an ego-trajectory-guided, geometry-aligned transformer for action-faithful occupancy world modeling. 
Its key design is to use the specified future trajectory to explicitly align each future query with relevant historical scene locations, thereby grounding full-scene forecasting in action-induced spatial correspondence. 
To this end, ReCast adopts a decoupled spatiotemporal encoder--decoder, where the encoder constructs a temporal memory exclusively from observed occupancy latents and the decoder jointly predicts the complete future sequence from scene-content-free, action-conditioned queries. 
Throughout the encoder and decoder, spatial deformable attention performs adaptive regional reasoning within each occupancy frame, while geometry-aligned temporal attention exchanges information across frames according to their relative ego transformations. 
This design separates action-induced spatial correspondence from non-rigid scene evolution. 
By combining explicit geometric alignment with adaptive regional interaction,
ReCast improves action faithfulness across both in-distribution and
OOD trajectories while preserving flexibility in full-scene
forecasting.

Experiments on Occ3D-nuScenes demonstrate that ReCast achieves competitive
forecasting performance against state-of-the-art occupancy world
models under the standard evaluation setting.
To isolate action faithfulness from the scene--trajectory correlations present
in standard benchmarks, we further introduce an evaluation protocol based on
controlled counterfactual scene--trajectory compositions.
By re-expressing each nominal future scene under alternative ego trajectories
spanning in-distribution, tail, and OOD actions, the protocol evaluates whether a model geometrically follows the specified ego
action under scene--trajectory compositions not jointly observed during
training.
Across the resulting \emph{CF-ID}, \emph{CF-Tail}, and \emph{CF-OOD} subsets,
ReCast consistently improves counterfactual forecasting performance, with particularly clear gains under OOD trajectories.

Our main contributions are summarized as follows:
\begin{itemize}
\item We expose action-faithfulness limitations of existing occupancy world
models that are obscured by scene--trajectory correlations in standard
benchmarks.

\item We propose ReCast, a geometry-aligned occupancy transformer that
grounds future prediction in action-induced spatial correspondence while
retaining flexibility for non-rigid scene evolution.

\item We introduce a counterfactual benchmark spanning in-distribution, tail,
and out-of-distribution trajectories, where ReCast demonstrates stronger
action faithfulness with competitive standard forecasting performance.
\end{itemize}

% Our main contributions are summarized as follows:
% \begin{itemize}
% \item We reveal that existing occupancy world models can remain biased by scene--trajectory correlations, leading to action-inconsistent forecasts even for in-distribution counterfactual actions and increasingly severe failures toward tail and out-of-distribution trajectories.

% \item We propose \textbf{ReCast}, a geometry-aligned occupancy transformer that explicitly grounds future prediction in action-induced spatial correspondence through relative ego transformations.

% \item We introduce a counterfactual action-faithfulness evaluation protocol spanning in-distribution, tail, and out-of-distribution actions, and show that ReCast achieves competitive standard forecasting performance with substantially stronger action faithfulness.
% \end{itemize}
\section{Related Work}
\label{2_related_work}
\subsection{4D Occupancy World Models}
Occupancy world models forecast the evolution of semantic 3D scenes from
historical observations.
OccWorld~\cite{zheng2024occworld} jointly generates future occupancy and ego
trajectories using an autoregressive token model.
Subsequent approaches have explored controllable generation by conditioning
occupancy on ego trajectories and driving actions.
Representative examples include DOME~\cite{gu2024dome},
Drive-OccWorld~\cite{yang2025driving},
DynamicCity~\cite{bian2025dynamiccity}, and
OccSora~\cite{wang2026occsora}.
OccLLaMA~\cite{wei2024occllama} further unifies occupancy, language, and action
within an autoregressive generative model.
Other methods focus on improving occupancy representation and sequence
modeling.
UniScene~\cite{li2025uniscene} introduces a unified generative representation
for driving scenes, RenderWorld~\cite{yan2025renderworld} combines
self-supervised occupancy labeling with compressed world modeling, and
Occ-LLM~\cite{xu2025occ} combines occupancy representations with large
language models for forecasting.
$I^2$-World~\cite{liao2025i2} improves long-horizon prediction through
structured occupancy tokenization and temporal modeling.

Despite being action-conditioned, most existing methods encode the ego
trajectory or driving action through implicit feature or token conditioning
rather than explicit geometric correspondence.
This can cause incomplete geometric adaptation of the predicted scene to
counterfactual trajectories, especially as the action departs from the
training distribution.
ReCast instead aligns future prediction with the geometric transformation
induced by the ego trajectory.
% Despite being action-conditioned, most existing methods treat the future ego trajectory primarily as a learned condition and infer its spatial effect from scene--trajectory correlations in the training data.
% As a result, their predictions may remain biased toward the trajectory associated with the observed scene under counterfactual scene--action compositions, even when the specified action is in distribution.
% This inconsistency can become further amplified for tail and out-of-distribution actions.
% In contrast, we explicitly target action faithfulness by structurally grounding future prediction in the geometric correspondence induced by the specified ego trajectory.
\begin{figure*}[hbt!]
    \centering
    \includegraphics[width=\linewidth]{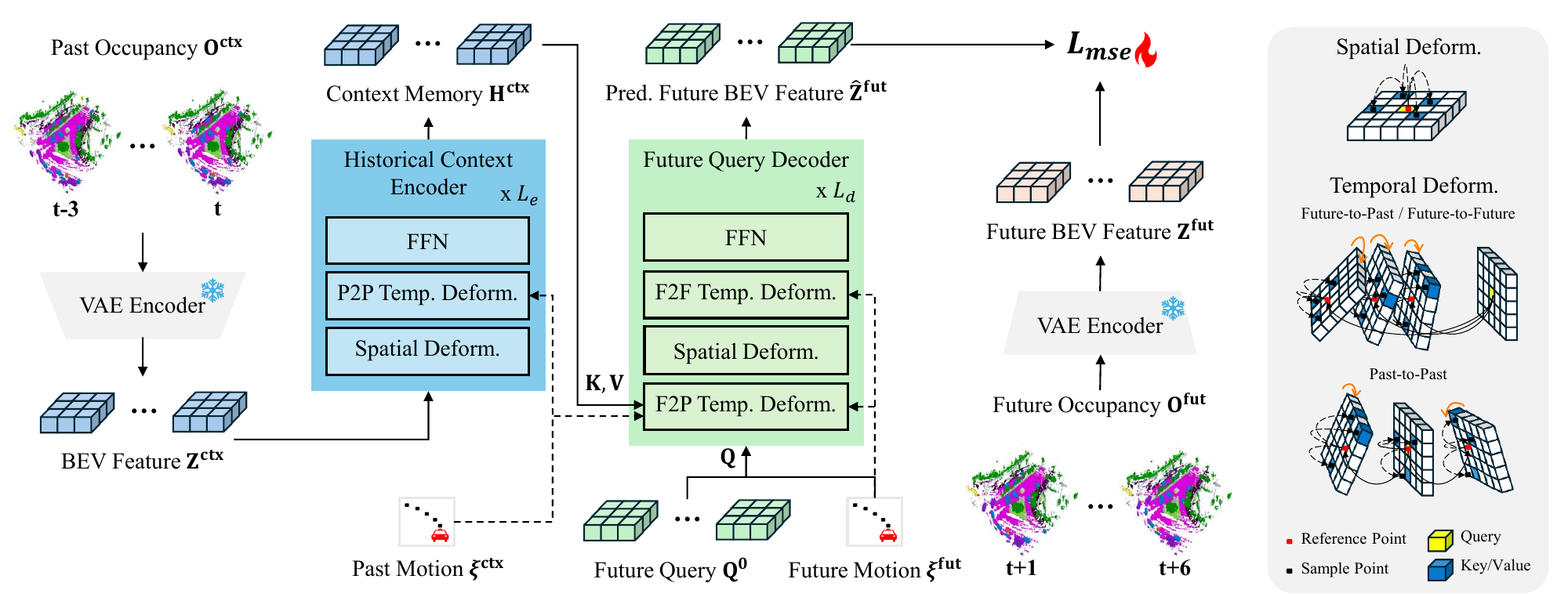}
    \caption{
Overall framework of the proposed ReCast.
Solid arrows indicate feature propagation, whereas dashed arrows indicate ego-motion inputs used for geometric alignment and future query conditioning.
For clarity, the input projections and positional encodings applied to
$\mathbf{Z}^{\mathrm{ctx}}$ and $\mathbf{Q}^{0}$, as well as the output
projection applied to $\hat{\mathbf{Z}}^{\mathrm{fut}}$, are omitted.
}
    \label{fig2}
    \vspace{-3.mm}
\end{figure*}
\subsection{Ego-Motion-Aware Spatiotemporal Modeling}
Explicit ego-motion modeling has been widely used to improve temporal
correspondence in BEV-based perception.
BEVFormer~\cite{li2024bevformer} recurrently aggregates historical BEV
features through temporal attention, while
BEVDet4D~\cite{huang2022bevdet4d} aligns adjacent-frame features before
fusion.
Similar geometric reasoning extends to future scene prediction.
ViDAR~\cite{yang2024visual} transforms historical representations into a 3D
geometric space for point-cloud forecasting, while
Drive-OccWorld~\cite{yang2025driving} combines geometric and temporal
modeling with flexible action conditioning.
D$^2$-World~\cite{zhang2024d} and
DFIT-OccWorld~\cite{zhang2024efficient} separate ego-motion-induced
static changes from dynamic evolution through pose transformation and voxel
flow, while COME~\cite{shi2026come} decouples scene evolution from viewpoint
changes using a scene-centric prediction.

Our distinction lies in how ego geometry is integrated into spatiotemporal
reasoning.
ReCast uses relative ego transformations to define temporal deformable
attention references throughout historical encoding and future decoding.
Rather than transforming occupancy content directly, it retrieves
geometrically corresponding evidence while adaptively modeling non-rigid scene
evolution.
Spatial deformable attention further aggregates regional context, enabling
action-grounded and flexible scene forecasting.
\section{Method}
\label{3_method}
Given four historical semantic occupancy grids,
$\mathbf{O}^{\mathrm{ctx}}=\{\mathbf{O}_{t-3},\ldots,\mathbf{O}_{t}\}$,
we jointly predict the next $T_f=6$ frames,
$\widehat{\mathbf{O}}^{\mathrm{fut}}
=\{\widehat{\mathbf{O}}_{t+1},\ldots,\widehat{\mathbf{O}}_{t+T_f}\}$,
conditioned on the historical and future ego-motion sequences
$\boldsymbol{\xi}^{\mathrm{ctx}}
=\{\boldsymbol{\xi}_{i}\}_{i=t-3}^{t}
\in\mathbb{R}^{4\times3}$
and
$\boldsymbol{\xi}^{\mathrm{fut}}
=\{\boldsymbol{\xi}_{t+\tau}\}_{\tau=1}^{T_f}
\in\mathbb{R}^{T_f\times3}$,
respectively.
Each element
$\boldsymbol{\xi}_{i}
=(\Delta x_i,\Delta y_i,\Delta\theta_i)$
denotes the relative planar pose of the ego vehicle at time $i$ with respect
to the last historical frame at time $t$, where
$\boldsymbol{\xi}_{t}=(0,0,0)$.
% where
% $(\Delta x_{t+\tau},\Delta y_{t+\tau})$ and
% $\Delta\theta_{t+\tau}$ represent the relative planar translation and yaw
% rotation, respectively.
Each occupancy grid
$\mathbf{O}_{i}\in\{0,\ldots,P-1\}^{X\times Y\times Z}$
represents a voxelized 3D scene with $P$ semantic classes, including empty space.
For efficient volumetric modeling, we adopt the pretrained Occ-VAE from
DOME~\cite{gu2024dome} and encode each grid as
$\mathbf{z}_{i}
=\mathcal{E}_{\mathrm{vae}}(\mathbf{O}_{i})
\in\mathbb{R}^{C_0\times H\times W}$,
where $(X,Y,Z)$ and $(H,W)$ denote the voxel and BEV latent resolutions,
respectively, and $C_0$ is the latent channel dimension.

ReCast takes the historical latent sequence
$\mathbf{Z}^{\mathrm{ctx}}
=\{\mathbf{z}_{t-3},\ldots,\mathbf{z}_{t}\}$
and future query grids
$\mathbf{Q}^{0}
=\{\mathbf{Q}^{0}_{\tau}\}_{\tau=1}^{T_f}
\in\mathbb{R}^{T_f\times C\times H\times W}$, where $C$ denotes the feature dimension.
ReCast jointly predicts the future latent sequence as:
\begin{equation}
\widehat{\mathbf{Z}}^{\mathrm{fut}}
=
\mathcal{F}
\left(
\mathbf{Z}^{\mathrm{ctx}},
\mathbf{Q}^{0},
\boldsymbol{\xi}^{\mathrm{ctx}},
\boldsymbol{\xi}^{\mathrm{fut}}
\right)
=
\left\{
\widehat{\mathbf{z}}_{t+1},
\ldots,
\widehat{\mathbf{z}}_{t+T_f}
\right\}.
\end{equation}
The frozen VAE decoder reconstructs
$\widehat{\mathbf{O}}^{\mathrm{fut}}
=\mathcal{D}_{\mathrm{vae}}(\widehat{\mathbf{Z}}^{\mathrm{fut}})$.
% The detailed architecture is described in the following subsection.

\subsection{ReCast: A Geometry-Aligned Transformer}
\label{sec:ReCast}
\subsubsection{Decoupled Past--Future Architecture}
ReCast decouples observed scene encoding from future prediction.
Joint processing may contaminate context features with uncertain future states
and bias them toward the historical layout.
We therefore construct memory exclusively from observed occupancy, from which
each future query retrieves horizon-specific evidence.

The model consists of a historical context encoder
$\mathcal{F}_{\mathrm{enc}}$ and a future query decoder
$\mathcal{F}_{\mathrm{dec}}$. The encoder first constructs the spatiotemporal
context memory
\begin{equation}
\mathbf{H}^{\mathrm{ctx}}
=
\mathcal{F}_{\mathrm{enc}}
\left(
\mathbf{Z}^{\mathrm{ctx}},
\boldsymbol{\xi}^{\mathrm{ctx}}
\right)
\in
\mathbb{R}^{4\times C\times H\times W},
\end{equation}
which is subsequently used by the decoder to predict the complete future latent
sequence:
\begin{equation}
\widehat{\mathbf{Z}}^{\mathrm{fut}}
=
\mathcal{F}_{\mathrm{dec}}
\left(
\mathbf{Q}^{0},
\mathbf{H}^{\mathrm{ctx}},
\boldsymbol{\xi}^{\mathrm{ctx}},
\boldsymbol{\xi}^{\mathrm{fut}}
\right).
\end{equation}

\subsubsection{Historical Context Encoder}

Each historical encoder block applies spatial deformable attention, 
ego-motion-aligned temporal deformable attention, and a feed-forward network.
The spatial module aggregates regional evidence within each frame, whereas the
temporal module aligns sampling references across historical frames using
relative ego transformations derived from
$\boldsymbol{\xi}^{\mathrm{ctx}}$.

For each historical frame $i\in\{t-3,\ldots,t\}$, the latent grid
$\mathbf{z}_{i}$ is first projected from $C_0$ to the Transformer dimension $C$ and
augmented with learnable spatial and frame-wise temporal embeddings to form the
initial frame representation $\mathbf{H}^{0}_{i}$.
% Both the spatial and temporal attention modules then operate through deformable
% sampling, differing primarily in how their geometric reference points are
% defined.
Let $\mathbf{H}^{\ell}_{i}$ denote the feature grid at encoder layer $\ell$.
For target position $p$ in frame $i$, the sampling set in source frame $j$ is
\begin{equation}
\label{eq:deformable_sampling}
\mathcal{P}_{j\leftarrow i}(p)
=
\left\{
\overline{p}_{j\leftarrow i}(p)
+
\Delta p^{(k)}_{j\leftarrow i}(p)
\right\}_{k=1}^{K},
\end{equation}
where $K$ denotes the number of sampling points for the corresponding
attention module, and $\Delta p^{(k)}_{j\leftarrow i}(p)$ is the learned
offset of its $k$-th sampling point. We use
$\mathcal{D}(\mathbf{Q},\mathbf{K},\mathbf{V};\mathcal{P})$ to denote
deformable attention with query, key, and value features
$\mathbf{Q}$, $\mathbf{K}$, and $\mathbf{V}$, respectively, and sampling set
$\mathcal{P}$.

Spatial attention uses $\mathbf{H}^{\ell}_{i}$ as query, key, and value, with
the query position itself as the reference:
\begin{equation}
\overline{p}^{\mathrm{sp}}(p)
=
p,
\qquad
\mathcal{P}^{\mathrm{sp}}(p)
=
\left\{
p+\Delta p^{\mathrm{sp},(k)}(p)
\right\}_{k=1}^{K_{\mathrm{sp}}}.
\end{equation}
\begin{equation}
\mathbf{H}^{\mathrm{sp},\ell}_{i}
=
\mathbf{H}^{\ell}_{i}
+
\mathcal{D}_{\mathrm{sp}}
\left(
\mathbf{H}^{\ell}_{i},
\mathbf{H}^{\ell}_{i},
\mathbf{H}^{\ell}_{i};
\mathcal{P}^{\mathrm{sp}}
\right).
\end{equation}

Temporal attention uses $\mathbf{H}^{\mathrm{sp},\ell}_{i}$ as the query and
$\mathcal{H}^{\ell}_{\mathrm{hist}}
=\{\mathbf{H}^{\mathrm{sp},\ell}_{j}\}_{j=t-3}^{t}$
as keys and values.
Let $\mathbf{T}_{j\leftarrow i}\in\mathrm{SE}(2)$ be the relative ego
transformation from frame $i$ to $j$, computed from
$\boldsymbol{\xi}^{\mathrm{ctx}}$.
The mapping $\mathcal{G}(p;\mathbf{T}_{j\leftarrow i})$ transforms a BEV
position from frame $i$ to $j$:
\begin{equation}
\overline{p}^{\mathrm{hist}}_{j\leftarrow i}(p)
=
\mathcal{G}
\left(
p;
\mathbf{T}_{j\leftarrow i}
\right).
\end{equation}
\begin{equation}
\mathcal{P}^{\mathrm{hist}}_{j\leftarrow i}(p)
=
\left\{
\overline{p}^{\mathrm{hist}}_{j\leftarrow i}(p)
+
\Delta p^{\mathrm{hist},(k)}_{j\leftarrow i}(p)
\right\}_{k=1}^{K_{\mathrm{hist}}}.
\end{equation}
For target frame $i$, the sampling sets across all source frames are
collectively denoted by
$\mathcal{P}^{\mathrm{hist}}_{i}
=\{\mathcal{P}^{\mathrm{hist}}_{j\leftarrow i}\}_{j=t-3}^{t}$.
\begin{equation}
\mathbf{H}^{\mathrm{tmp},\ell}_{i}
=
\mathbf{H}^{\mathrm{sp},\ell}_{i}
+
\mathcal{D}_{\mathrm{hist}}
\left(
\mathbf{H}^{\mathrm{sp},\ell}_{i},
\mathcal{H}^{\ell}_{\mathrm{hist}},
\mathcal{H}^{\ell}_{\mathrm{hist}};
\mathcal{P}^{\mathrm{hist}}_{i}
\right).
\end{equation}
\begin{equation}
\mathbf{H}^{\ell+1}_{i}
=
\mathbf{H}^{\mathrm{tmp},\ell}_{i}
+
\operatorname{FFN}_{\mathrm{enc}}
\left(
\mathbf{H}^{\mathrm{tmp},\ell}_{i}
\right).
\end{equation}

The final frame-wise features form the context memory
\begin{equation}
\mathbf{H}^{\mathrm{ctx}}
=
\operatorname{Stack}
\left(
\mathbf{H}^{L_{\mathrm{e}}}_{t-3},
\ldots,
\mathbf{H}^{L_{\mathrm{e}}}_{t}
\right)
\in
\mathbb{R}^{4\times C\times H\times W}.
\end{equation}

\subsubsection{Future Query Decoder}

For each horizon $\tau$, the scene-content-free query
$\mathbf{Q}^{0}_{\tau}$ combines a shared mask token with learnable spatial
and horizon embeddings and an action embedding from
$\boldsymbol{\xi}_{t+\tau}$.
Each decoder block applies geometry-aligned future-to-past cross-attention,
spatial self-attention, causal preceding-frame attention, and an FFN.

Historical and future poses define the relative transformations
$\mathbf{T}_{i\leftarrow t+\tau}$ and
$\mathbf{T}_{t+\tau-1\leftarrow t+\tau}$. For a position $p$ in future frame $t+\tau$, its reference point in
historical frame $i$ is
\begin{equation}
\overline{p}^{\mathrm{ctx}}_{i\leftarrow\tau}(p)
=
\mathcal{G}
\left(
p;
\mathbf{T}_{i\leftarrow t+\tau}
\right).
\end{equation}
Following Eq.~\eqref{eq:deformable_sampling}, let
$\mathcal{P}^{\mathrm{ctx}}_{\tau}
=\{\mathcal{P}^{\mathrm{ctx}}_{i\leftarrow\tau}\}_{i=t-3}^{t}$, with
$\mathbf{Q}^{\ell}_{\tau}$ as query and $\mathbf{H}^{\mathrm{ctx}}$ as keys
and values.
\begin{equation}
\widetilde{\mathbf{Q}}^{\ell}_{\tau}
=
\mathbf{Q}^{\ell}_{\tau}
+
\mathcal{D}_{\mathrm{ctx}}
\left(
\mathbf{Q}^{\ell}_{\tau},
\mathbf{H}^{\mathrm{ctx}},
\mathbf{H}^{\mathrm{ctx}};
\mathcal{P}^{\mathrm{ctx}}_{\tau}
\right).
\end{equation}
If none of the deformable samples provides valid historical evidence, the
attention output is replaced by a learnable invalid-context token.

Spatial attention refines the retrieved evidence using the query position as
its reference:
\begin{equation}
\mathcal{P}^{\mathrm{sp}}_{\tau}(p)
=
\left\{
p+\Delta p^{\mathrm{sp},(k)}_{\tau}(p)
\right\}_{k=1}^{K_{\mathrm{sp}}}.
\end{equation}
\begin{equation}
\mathbf{Q}^{\mathrm{sp},\ell}_{\tau}
=
\widetilde{\mathbf{Q}}^{\ell}_{\tau}
+
\mathcal{D}_{\mathrm{sp}}
\left(
\widetilde{\mathbf{Q}}^{\ell}_{\tau},
\widetilde{\mathbf{Q}}^{\ell}_{\tau},
\widetilde{\mathbf{Q}}^{\ell}_{\tau};
\mathcal{P}^{\mathrm{sp}}_{\tau}
\right).
\end{equation}

Causal temporal attention retrieves from the immediately preceding future
frame. For $\tau>1$, the aligned reference is
\begin{equation}
\overline{p}^{\mathrm{fut}}_{\tau-1\leftarrow\tau}(p)
=
\mathcal{G}
\left(
p;
\mathbf{T}_{t+\tau-1\leftarrow t+\tau}
\right).
\end{equation}
Following Eq.~\eqref{eq:deformable_sampling}, let
$\mathcal{P}^{\mathrm{fut}}_{\tau-1\leftarrow\tau}$ denote its sampling set.
\begin{equation}
\mathbf{Q}^{\mathrm{tmp},\ell}_{\tau}
=
\mathbf{Q}^{\mathrm{sp},\ell}_{\tau}
+
\mathcal{D}_{\mathrm{fut}}
\left(
\mathbf{Q}^{\mathrm{sp},\ell}_{\tau},
\mathbf{Q}^{\mathrm{sp},\ell}_{\tau-1},
\mathbf{Q}^{\mathrm{sp},\ell}_{\tau-1};
\mathcal{P}^{\mathrm{fut}}_{\tau-1\leftarrow\tau}
\right).
\end{equation}
For $\tau=1$, we set
$\mathbf{Q}^{\mathrm{tmp},\ell}_{1}
=\mathbf{Q}^{\mathrm{sp},\ell}_{1}$.

\begin{equation}
\mathbf{Q}^{\ell+1}_{\tau}
=
\mathbf{Q}^{\mathrm{tmp},\ell}_{\tau}
+
\operatorname{FFN}_{\mathrm{fut}}
\left(
\mathbf{Q}^{\mathrm{tmp},\ell}_{\tau}
\right).
\end{equation}

After the final decoder layer, the features are projected to dimension $C_0$,
yielding $\widehat{\mathbf{Z}}^{\mathrm{fut}}$.

\subsubsection{Loss}
Training minimizes future-latent MSE:
\begin{equation}
\mathcal{L}_{\mathrm{mse}}
=
\frac{1}{T_f C_0 H W}
\sum_{\tau=1}^{T_f}
\left\|
\widehat{\mathbf{z}}_{t+\tau}
-
\mathbf{z}_{t+\tau}
\right\|_{2}^{2}.
\end{equation}

\subsection{Action-Faithfulness Evaluation Protocol}
\label{sec:action_faithfulness_protocol}

\begin{figure}[hbt!]
    \centering
    \includegraphics[width=\linewidth]{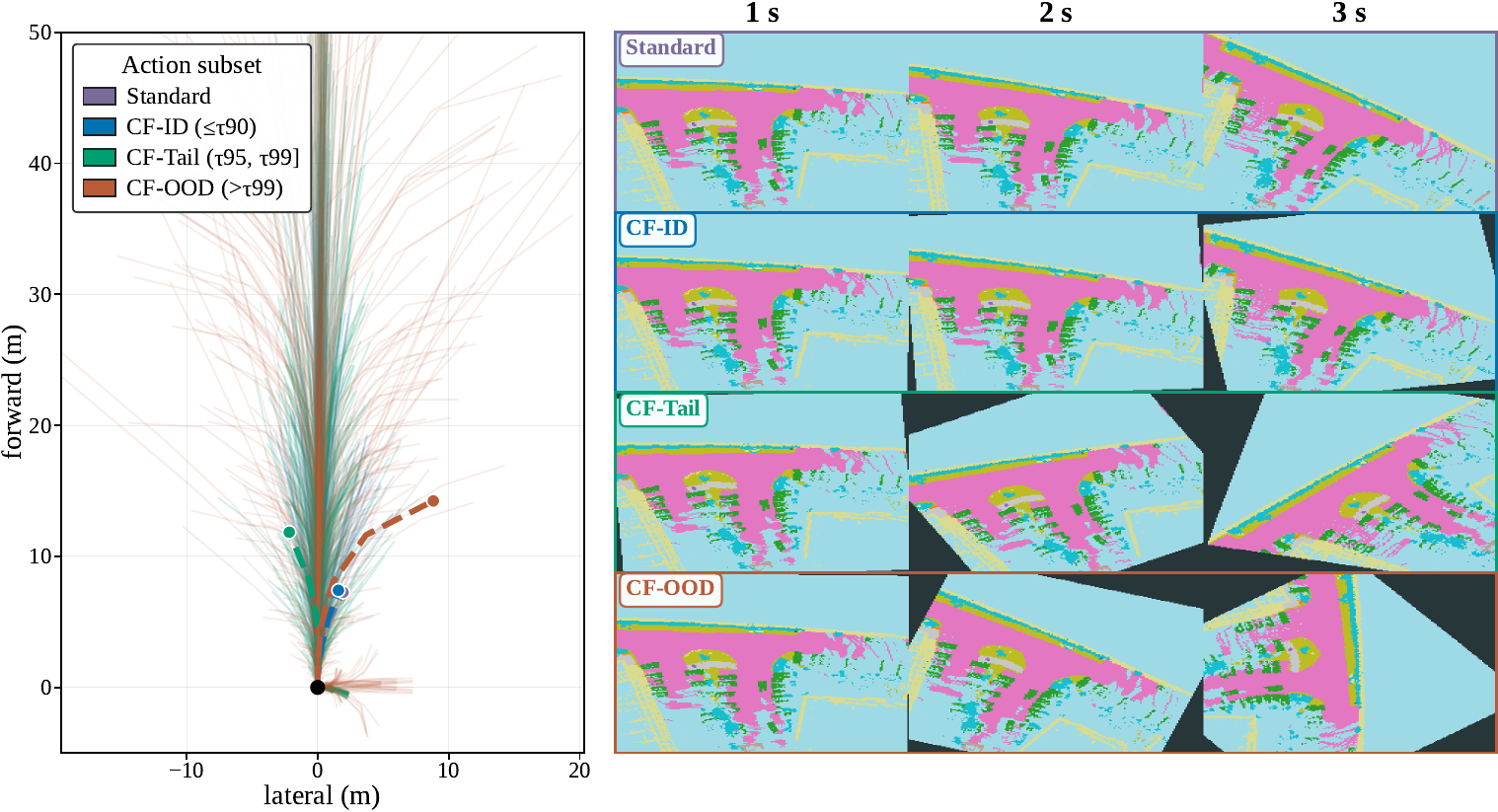}
    \caption{
    Left: standard and counterfactual future ego trajectories grouped by their
    density under the training action distribution.
    Right: representative occupancy targets at 1, 2, and 3 seconds for
    \emph{Standard}, \emph{CF-ID}, \emph{CF-Tail}, and \emph{CF-OOD}.
    }
    \label{fig3}
    \vspace{-3.mm}
\end{figure}

Standard validation trajectories largely follow the action distribution observed
during training and may therefore fail to distinguish models that faithfully
respond to the specified future ego motion from those that exploit observed
scene--trajectory correlations.
To evaluate this property, we construct a fixed, model-independent
counterfactual benchmark from 150 nuScenes validation scenes.
For each scene, we generate three counterfactual trajectories in each of three
action-density regimes, yielding 450 samples for \emph{CF-ID},
\emph{CF-Tail}, and \emph{CF-OOD}, respectively.
Together with the \emph{Standard} samples, these regimes evaluate
forecasting under nominal, common counterfactual, rare counterfactual, and
OOD counterfactual ego motions, as illustrated in
Fig.~\ref{fig3}.

\subsubsection{Action density and regimes}
We represent each future ego trajectory by concatenating its cumulative planar
translation and yaw over all prediction horizons:
\begin{equation}
\mathbf{a}
=
\left[
x_{t+1},y_{t+1},\theta_{t+1},
\ldots,
x_{t+T_f},y_{t+T_f},\theta_{t+T_f}
\right]^{\top}
\in
\mathbb{R}^{3T_f},
\end{equation}
where all poses are expressed relative to the last observed ego frame.
For $T_f=6$, this yields an 18-dimensional descriptor.
We standardize each dimension using statistics computed exclusively from the
training trajectories.
% New rows: Rigid Propagation, Standard 4,669-window corrected-nearest-support evaluation.
% Standard Rigid uses boundary-corrected nearest support; the archived CF Rigid result uses the prior validated CF sampler.
% CF GT: corrected-target rigid baseline from previously supplied cf1350 metrics; camera CF Change values supplied subsequently.
% Dynamic IoU uses CF GT dynamic ROI; not the full-grid dynamic-class IoU.
% GT CF: existing checkpoints; corrected targets, masks and ReCast action inference.
% Camera CF: corrected Full/Change with mIoU and occupancy IoU for all evaluated Camera models.
% Standard camera and prior-model standard metrics retained from submitted manuscript.
% Requires \usepackage{xcolor} in the preamble.
\begin{table*}[hbt!]
\centering
% \setlength{\tabcolsep}{3.pt}
% \renewcommand{\arraystretch}{1.08}
% \scriptsize
\resizebox{\linewidth}{!}{
\begin{tabular}{llcccccccccc}
\toprule
\multirow{3}{*}{Method}
&
\multirow{3}{*}{Input}
&
\multicolumn{4}{c}{Standard Forecasting (mIoU / IoU)}
&
\multicolumn{6}{c}{Action-Faithfulness (mIoU / IoU)}
\\
\cmidrule(lr){3-6}
\cmidrule(lr){7-12}
&
&
\multirow{2}{*}{1s}
&
\multirow{2}{*}{2s}
&
\multirow{2}{*}{3s}
&
\multirow{2}{*}{Avg.}
&
\multicolumn{2}{c}{CF-ID}
&
\multicolumn{2}{c}{CF-Tail}
&
\multicolumn{2}{c}{CF-OOD}
\\
\cmidrule(lr){7-8}
\cmidrule(lr){9-10}
\cmidrule(lr){11-12}
&
&
&
&
&
&
Full
&
Change
&
Full
&
Change
&
Full
&
Change
\\
\midrule

% OccWorld~\cite{zheng2024occworld}
% & 3D-Occ
% & 25.78 / 34.63
% & 15.14 / 25.07
% & 10.51 / 20.18
% & 17.14 / 26.63
% & -- & --
% & -- & --
% & -- & --
% \\

DFIT-OccWorld~\cite{zhang2024efficient}
& 3D-Occ
& 31.68 / 40.28
& 21.29 / 31.24
& 15.18 / 25.29
& 22.71 / 32.27
& -- / --
& -- / --
& -- / --
& -- / --
& -- / --
& -- / --
\\

DOME~\cite{gu2024dome}
& 3D-Occ
& 35.11 / 43.99
& 25.89 / 35.36
& 20.29 / 29.74
& 27.10 / 36.36
& 23.15 / 32.94
& 20.64 / 30.10
& 11.78 / 20.76
& 7.58 / 15.58
& 7.91 / 15.91
& 4.30 / 12.02
\\

UniScene~\cite{li2025uniscene}
& 3D-Occ
& 35.37 / 38.34
& 29.59 / 32.70
& 25.08 / 29.09
& 30.01 / 33.38
& -- / --
& -- / --
& -- / --
& -- / --
& -- / --
& -- / --
\\

COME~\cite{shi2026come}
& 3D-Occ
& 42.75 / 50.57
& 32.97 / 43.47
& 26.98 / 38.36
& 34.23 / 44.13
& 31.83 / 38.09
& 35.27 / 40.48
& 21.29 / 28.19
& 20.25 / 26.43
& 15.97 / 23.79
& 13.41 / 21.38
\\

$I^2$-World~\cite{liao2025i2}
& 3D-Occ
& \textbf{47.62} / \textbf{54.29}
& \textbf{38.58} / \textbf{49.43}
& \textbf{32.98} / \textbf{45.69}
& \textbf{39.73} / \textbf{49.80}
& \underline{34.79} / \underline{40.81}
& \underline{35.40} / \underline{41.35}
& \underline{32.20} / \underline{38.46}
& \underline{35.38} / \underline{42.39}
& \underline{23.14} / \underline{29.14}
& \underline{24.94} / \underline{31.27}
\\

\textbf{ReCast}
& 3D-Occ
& \underline{43.94} / \underline{51.51}
& \underline{36.53} / \underline{46.79}
& \underline{31.85} / \underline{43.34}
& \underline{37.44} / \underline{47.21}
& \textbf{37.79} / \textbf{46.57}
& \textbf{39.91} / \textbf{48.45}
& \textbf{34.64} / \textbf{43.69}
& \textbf{38.48} / \textbf{48.34}
& \textbf{31.13} / \textbf{40.33}
& \textbf{35.54} / \textbf{45.82}
\\

\midrule
\textcolor{gray}{Rigid Propagation$^\ddagger$}
& \textcolor{gray}{3D-Occ}
& \textcolor{gray}{40.97} / \textcolor{gray}{50.45}
& \textcolor{gray}{32.52} / \textcolor{gray}{41.53}
& \textcolor{gray}{27.77} / \textcolor{gray}{36.25}
& \textcolor{gray}{33.75} / \textcolor{gray}{42.75}
& \textcolor{gray}{36.24} / \textcolor{gray}{45.29}
& \textcolor{gray}{39.51} / \textcolor{gray}{47.95}
& \textcolor{gray}{35.58} / \textcolor{gray}{44.46}
& \textcolor{gray}{41.37} / \textcolor{gray}{50.59}
& \textcolor{gray}{35.85} / \textcolor{gray}{44.63}
& \textcolor{gray}{42.54} / \textcolor{gray}{51.66}
\\

\midrule

% OccWorld~\cite{zheng2024occworld}
% & Camera
% & 10.97 / 20.68
% & 15.16 / 24.81
% & 10.19 / 20.09
% & 7.53 / 17.14
% & -- & --
% & -- & --
% & -- & --
% \\

DOME~\cite{gu2024dome}
& Camera
& 17.79 / 26.39
& 14.23 / 23.20
& 11.58 / 20.42
& 14.53 / 23.33
& 12.36 / 20.93
& 11.92 / 18.58
& 7.28 / 15.74
& 5.68 / 12.71
& 5.25 / 13.03
& 3.60 / 10.72
\\

COME$^\dagger$~\cite{shi2026come}
& Camera
& 17.28 / 24.19
& 14.47 / 22.02
& 12.25 / 20.37
& 14.67 / 22.19
& 14.81 / 22.07
& 16.45 / 21.02
& \underline{11.06} / 18.19
& 11.15 / 16.35
& 8.67 / 15.84
& 8.13 / 14.32
\\

$I^2$-World~\cite{liao2025i2}
& Camera
& \textbf{21.67} / \textbf{30.55}
& \textbf{18.78} / \textbf{28.76}
& \textbf{16.47} / \textbf{26.99}
& \textbf{18.97} / \textbf{28.77}
& \underline{15.50} / \underline{23.30}
& \underline{16.99} / \textbf{21.98}
& \textbf{15.10} / \textbf{22.88}
& \textbf{17.54} / \textbf{23.50}
& \underline{11.23} / \underline{18.07}
& \underline{12.96} / \underline{18.42}
\\

\textbf{ReCast}
& Camera
& \underline{20.21} / \underline{28.02}
& \underline{17.36} / \underline{25.84}
& \underline{14.99} / \underline{23.92}
& \underline{17.52} / \underline{25.93}
& \textbf{15.76} / \textbf{23.35}
& \textbf{17.12} / \underline{21.51}
& \textbf{15.10} / \underline{22.82}
& \underline{17.19} / \underline{22.83}
& \textbf{14.05} / \textbf{21.82}
& \textbf{16.69} / \textbf{23.04}
\\

\midrule
\textcolor{gray}{Rigid Propagation$^\ddagger$}
& \textcolor{gray}{Camera}
& \textcolor{gray}{18.47} / \textcolor{gray}{27.52}
& \textcolor{gray}{15.65} / \textcolor{gray}{24.30}
& \textcolor{gray}{13.63} / \textcolor{gray}{21.66}
& \textcolor{gray}{15.92} / \textcolor{gray}{24.49}
& \textcolor{gray}{14.47} / \textcolor{gray}{22.85}
& \textcolor{gray}{17.30} / \textcolor{gray}{22.36}
& \textcolor{gray}{14.37} / \textcolor{gray}{22.71}
& \textcolor{gray}{17.94} / \textcolor{gray}{23.96}
& \textcolor{gray}{14.31} / \textcolor{gray}{22.60}
& \textcolor{gray}{18.67} / \textcolor{gray}{25.44}
\\

\bottomrule
\end{tabular}
}

\caption{
Quantitative comparison on Occ3D-nuScenes under standard and
counterfactual forecasting.
Camera-input methods share STCOcc~\cite{liao2025stcocc}
historical predictions; $^\dagger$ marks reproduced standard results.
Counterfactual results average six horizons.
\emph{Full} covers all valid voxels, while \emph{Change} covers
voxels with different nominal and counterfactual semantic labels.
$^\ddagger$ denotes training-free rigid propagation.
Bold and underlined scores indicate the top two learned methods,
excluding the gray rigid baseline.
}

\label{tab:quantitative_comparison}
\end{table*}

% CF Dynamic mIoU: full-grid dynamic semantic classes. Dynamic IoU: CF GT dynamic ROI (recall-like).
% CF Unseen: unseen_geometric_corrected (geometric visibility proxy, not physical occlusion).
% Standard Unseen: corrected geometric visibility mask; archived predictions aggregated on 4,669 windows.
\begin{table*}[hbt!]
\centering
\setlength{\tabcolsep}{15.0pt}
% \renewcommand{\arraystretch}{1.08}
% \scriptsize
\resizebox{\linewidth}{!}{
\begin{tabular}{lcccccccc}
\toprule
\multirow{2}{*}{Method}
&
\multicolumn{2}{c}{Standard}
&
\multicolumn{2}{c}{CF-ID}
&
\multicolumn{2}{c}{CF-Tail}
&
\multicolumn{2}{c}{CF-OOD}
\\
\cmidrule(lr){2-3}
\cmidrule(lr){4-5}
\cmidrule(lr){6-7}
\cmidrule(lr){8-9}
&
Dynamic & Unseen
& Dynamic & Unseen
& Dynamic & Unseen
& Dynamic & Unseen
\\
\midrule

COME~\cite{shi2026come}
& 32.17 / \textbf{57.43}
& 5.42 / 12.28
& 29.73 / \underline{53.36}
& 5.14 / 12.89
& 21.02 / 39.44
& 3.10 / 10.47
& 15.61 / 31.23
& 2.61 / 9.13
\\

$I^2$-World~\cite{liao2025i2}
& \textbf{35.46} / \underline{55.95}
& \textbf{11.16} / \textbf{21.77}
& \underline{31.73} / 52.12
& \underline{8.90} / \underline{16.24}
& \underline{29.85} / \underline{49.44}
& \underline{8.06} / \underline{15.38}
& \underline{21.64} / \underline{37.47}
& \underline{5.69} / \underline{11.79}
\\

\textbf{ReCast}
& \underline{34.86} / 55.86
& \underline{10.07} / \underline{20.00}
& \textbf{33.37} / \textbf{53.76}
& \textbf{9.61} / \textbf{19.96}
& \textbf{30.58} / \textbf{50.42}
& \textbf{8.43} / \textbf{18.28}
& \textbf{27.75} / \textbf{46.57}
& \textbf{6.88} / \textbf{15.66}
\\

\midrule
\textcolor{gray}{Rigid Propagation$^\ddagger$}
& \textcolor{gray}{30.25} / \textcolor{gray}{47.37}
& \textcolor{gray}{0.00} / \textcolor{gray}{0.00}
& \textcolor{gray}{29.41} / \textcolor{gray}{46.08}
& \textcolor{gray}{0.00} / \textcolor{gray}{0.00}
& \textcolor{gray}{28.99} / \textcolor{gray}{45.33}
& \textcolor{gray}{0.00} / \textcolor{gray}{0.00}
& \textcolor{gray}{29.26} / \textcolor{gray}{45.46}
& \textcolor{gray}{0.00} / \textcolor{gray}{0.00}
\\

\bottomrule
\end{tabular}
}

\caption{
Performance breakdown on dynamic classes and geometric
out-of-support regions under standard and counterfactual forecasting.
Results are six-horizon averages (mIoU / occupancy IoU, \%).
Dynamic mIoU covers eight classes under the legacy
class-absence convention; occupancy IoU uses the GT dynamic ROI.
Unseen denotes voxels outside the geometric support of all four
historical occupancy grids, without patch expansion or
sensor-occlusion interpretation; mIoU uses union-supported classes.
$^\ddagger$ denotes training-free rigid propagation.
Bold and underlined scores indicate the top two learned methods,
excluding the gray rigid baseline.
}

\label{tab:regional_breakdown}
\end{table*}
We then estimate action rarity using the average distance to the
$K_{\mathrm{nn}}$ nearest standardized training descriptors:
\begin{equation}
s(\mathbf{a})
=
\frac{1}{K_{\mathrm{nn}}}
\sum_{\widetilde{\mathbf{a}}_{j}
\in
\mathcal{N}_{K_{\mathrm{nn}}}(\widetilde{\mathbf{a}})}
\left\|
\widetilde{\mathbf{a}}
-
\widetilde{\mathbf{a}}_{j}
\right\|_{2}.
\end{equation}
Using the 90th, 95th, and 99th percentiles of the training-score distribution,
denoted by $\tau_{90}$, $\tau_{95}$, and $\tau_{99}$, we define
\begin{equation}
\begin{aligned}
\mathrm{CF\mbox{-}ID}:&\quad s(\mathbf{a}) \leq \tau_{90},\\
\mathrm{CF\mbox{-}Tail}:&\quad \tau_{95} < s(\mathbf{a}) \leq \tau_{99},\\
\mathrm{CF\mbox{-}OOD}:&\quad s(\mathbf{a}) > \tau_{99}.
\end{aligned}
\end{equation}
The density reference and thresholds are computed from training trajectories
and fixed for all models and samples.
% A small number of retained CF-Tail samples (7 of 450)
% fall outside the strict density interval due to
% relaxed proposal completion, as detailed in the Appendix.

\subsubsection{Counterfactual generation}
For each selected nominal sample, we generate counterfactual
trajectories by perturbing speed, curvature, or both while
keeping the observed history fixed.
We screen candidate trajectories using predefined geometric
and kinematic constraints and assign them to the density
regimes above, obtaining three counterfactual actions per
regime.

Given an accepted trajectory, we construct its occupancy target by
re-expressing the nominal future occupancy in the counterfactual
ego frame. Let $\mathcal{W}(\mathbf{O};\mathbf{T})$ denote semantic voxel
warping of occupancy $\mathbf{O}$ under transformation $\mathbf{T}$:
\begin{equation}
\mathbf{O}^{\mathrm{cf}}_{t+\tau}
=
\mathcal{W}
\left(
\mathbf{O}_{t+\tau};
\mathbf{T}^{\mathrm{cf}\leftarrow\mathrm{nom}}_{t+\tau}
\right),
\end{equation}
where $\mathbf{T}^{\mathrm{cf}\leftarrow\mathrm{nom}}_{t+\tau}$ maps
coordinates from the nominal future ego frame to the counterfactual future ego
frame. Thus, each paired sample preserves the observed history and nominal
world evolution while changing only the future ego motion used to express the
target. This isolates the model's geometric response to the specified action
without simulating counterfactual reactions of other scene agents.
Validity masks are transformed consistently with the
counterfactual targets. We independently verify the
corrected geometric transformations and assess
round-trip consistency and valid-region coverage.
All 1,350 samples are retained for evaluation, with
detailed quality-control and kinematic diagnostics
reported in the Appendix.
\section{Experiment}

\subsection{Experimental Settings}
We conduct experiments on Occ3D-nuScenes~\cite{tian2023occ3d}, which provides 18-class semantic occupancy annotations for nuScenes~\cite{caesar2020nuscenes}. The dataset contains 700 training, 150 validation, and 150 test sequences over a $200\times200\times16$ voxel grid covering
$[-40,40]$\,m, $[-40,40]$\,m, and $[-1,5.4]$\,m
at a $0.4$\,m resolution. Following prior work, the main standard comparison reports geometric IoU and
semantic mIoU at 1, 2, and 3 seconds, averaged across the three horizons. To evaluate action faithfulness, we additionally report six-horizon average IoU and mIoU on the \emph{CF-ID}, \emph{CF-Tail}, and \emph{CF-OOD} subsets defined by our counterfactual evaluation protocol. Metrics are computed only over voxels that remain valid after counterfactual transformation.

\begin{figure*}[hbt!]
    \centering
    \includegraphics[width=\linewidth]{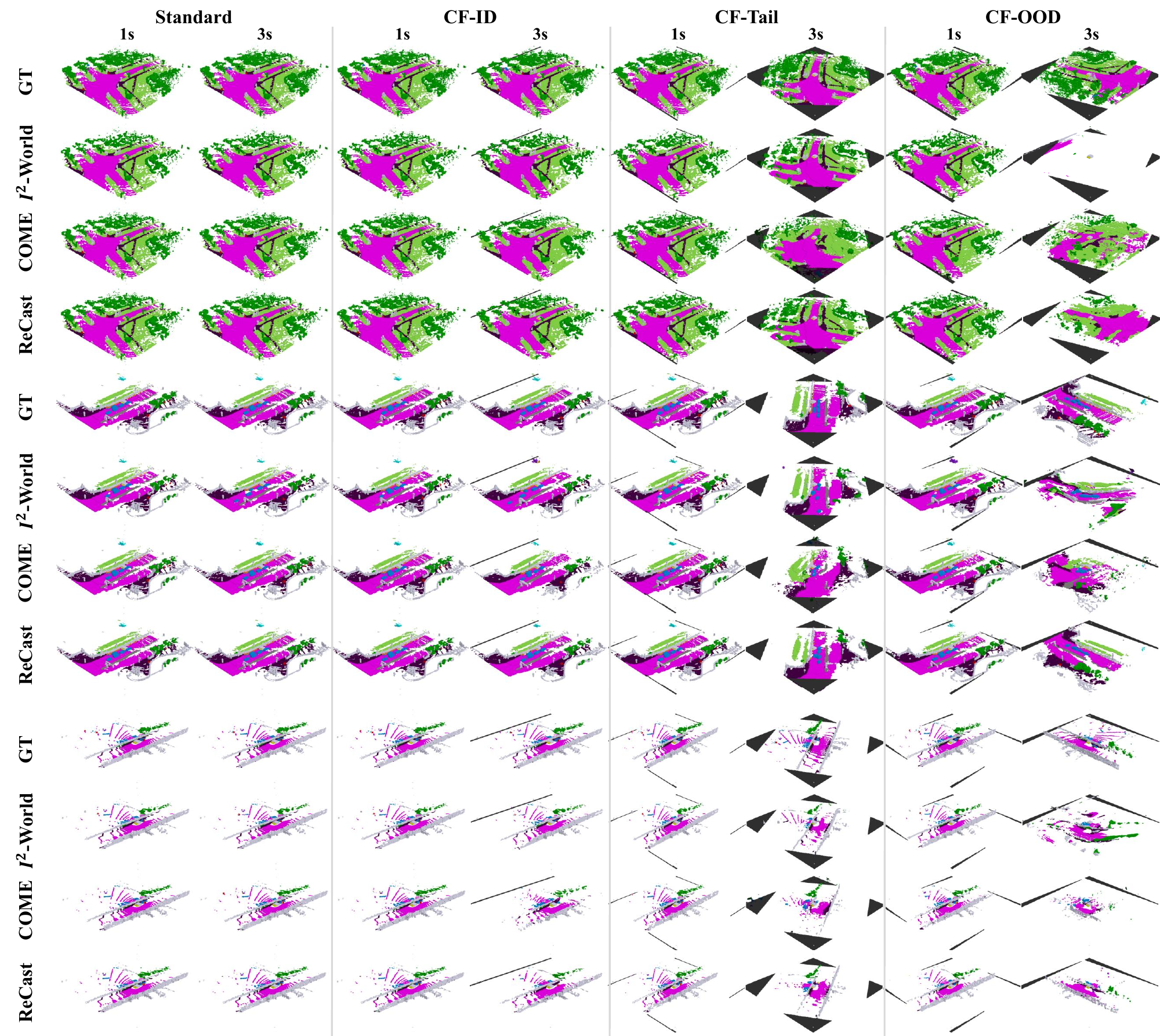}
    \caption{
Qualitative comparison among COME, $I^2$-World,
and ReCast under standard and counterfactual trajectories
at 1\,s and 3\,s.
Black regions indicate areas outside the valid occupancy range after
counterfactual transformation.
}
    \label{fig4}
    \vspace{-2.mm}
\end{figure*}

\begin{figure}[t]
    \centering
    \includegraphics[width=\linewidth]{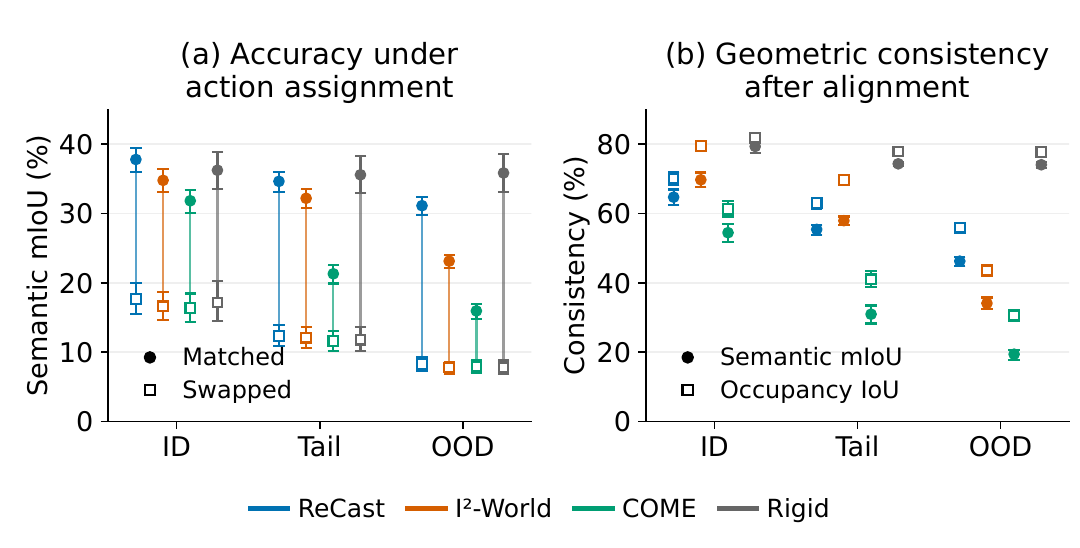}
\caption{
\textbf{Action intervention analysis.}
(a) Matched vs.\ swapped-action accuracy against fixed ground truth.
(b) Geometric consistency of coordinate-aligned predictions
(semantic mIoU / occupancy IoU).
Results average six horizons over 2,700 action swaps.
Error bars show 95\% scene-bootstrap confidence intervals.
Semantic mIoU excludes empty and uses union-supported classes.
}
    \label{fig5}
    \vspace{-3.mm}
\end{figure}

\subsection{Quantitative Evaluation}

Table~\ref{tab:quantitative_comparison} compares ReCast with recent occupancy
world models under both the standard forecasting protocol and the proposed
action-faithfulness benchmark.
Under the standard protocol, ReCast achieves competitive performance for both
3D occupancy and camera inputs, although $I^2$-World attains higher overall
forecasting accuracy.
This ranking changes under counterfactual evaluation, where the model must
adapt its prediction to a substituted ego trajectory.
Among directly action-conditioned learned models, ReCast
achieves the highest full-region mIoU across all
counterfactual regimes.
With 3D occupancy input, ReCast outperforms $I^2$-World by
$3.01$, $2.44$, and $7.99$ mIoU on CF-ID, CF-Tail, and CF-OOD, respectively.
With camera input, ReCast achieves a clear gain on CF-OOD
(+2.82 mIoU), while showing comparable performance on
CF-ID and CF-Tail.
The stronger gains with 3D occupancy input suggest that clean geometric
observations better expose the benefit of explicit alignment, whereas errors
in the camera-derived occupancy predictions may weaken the underlying spatial
correspondence.
These results show that strong performance under the recorded trajectory does
not necessarily translate into faithful adaptation to alternative ego actions.

Since our counterfactual targets are constructed by
geometrically re-expressing nominal future occupancy,
a simple rigid transformation may achieve high scores
without explicitly modeling future scene evolution.
To examine this potential shortcut, we introduce a
training-free rigid propagation baseline that transforms
the last observed occupancy into each specified future
ego frame, without predicting scene dynamics.
Rigid propagation indeed performs strongly under CF-Tail
and CF-OOD, achieving 35.58 and 35.85 mIoU,
surpassing ReCast's 34.64 and 31.13.
However, ReCast outperforms rigid propagation under
standard forecasting (37.44 vs.\ 33.75 mIoU) and CF-ID
(37.79 vs.\ 36.24 mIoU).
Moreover, ReCast predicts occupancy in strictly
out-of-support regions where rigid propagation cannot,
as further shown in Table~\ref{tab:regional_breakdown}.
These results indicate that ReCast goes beyond
simply transporting historical occupancy, while
highlighting the strength of geometric propagation
under large counterfactual action shifts.

Table~\ref{tab:regional_breakdown} further analyzes
forecasting performance on dynamic semantic classes
and strictly unsupported geometric regions.
Under standard forecasting, $I^2$-World
outperforms ReCast in both regions.
Under counterfactual evaluation, ReCast improves
dynamic-class mIoU over $I^2$-World by $1.64$,
$0.74$, and $6.11$ points on CF-ID, CF-Tail, and
CF-OOD, respectively.
For Unseen regions, ReCast achieves mIoU
gains of $0.71$, $0.37$, and $1.19$ points,
and occupancy IoU gains of $3.72$, $2.90$, and
$3.87$ points, respectively.
The semantic mIoU improvements are statistically
significant only under CF-OOD, while occupancy IoU
improves across all three regimes.
These results suggest that ReCast's counterfactual
improvements over learned baselines are not limited
to full-scene averages, but also extend to dynamic
semantics and geometrically unsupported regions,
with the clearest gains under CF-OOD.

\subsection{Qualitative Evaluation}
Figure~\ref{fig4} qualitatively compares ReCast with COME and $I^2$-World under
progressively larger counterfactual action shifts.
Although the mIoU improvements are consistent across all
counterfactual settings, the visual differences are relatively subtle under
CF-ID and CF-Tail, where the compared models generally produce plausible
future occupancy despite the substituted trajectories.
The distinction becomes more evident under CF-OOD, particularly at the
3-second horizon, where ReCast remains better aligned with the corresponding
ground truth.
This suggests that explicit geometric alignment becomes more important as the
specified trajectory departs further from the scene--trajectory combinations
observed during training.
These observations suggest that visual inspection alone is
insufficient to explain the quantitative improvements,
motivating the following action-intervention analysis.

Figure~\ref{fig5} investigates how occupancy forecasts respond to
counterfactual action interventions.
Keeping the observed scene and evaluation target fixed,
we replace each matched counterfactual action with another
action from the same density regime.
As shown in Fig.~\ref{fig5}(a), all learned models exhibit
lower accuracy under swapped actions, indicating that
they respond to the specified ego trajectory rather than
ignoring it entirely.
Notably, ReCast achieves the highest matched-action accuracy
and the largest accuracy drop under action swapping among
learned models across all three regimes.
These results suggest that ReCast combines more accurate
counterfactual forecasting with stronger sensitivity to
the specified action.
We further examine whether such action-dependent predictions
remain geometrically consistent across different trajectories.

Figure~\ref{fig5}(b) further evaluates geometric consistency.
For each pair of actions within the same regime, we
generate separate forecasts, align them to a common
coordinate frame using their relative ego transformations,
and measure their semantic and occupancy agreement.
Although $I^2$-World achieves higher consistency under
CF-ID and CF-Tail, ReCast exhibits a smaller decline
from CF-ID to CF-OOD (64.72 to 46.26 vs.\
69.73 to 34.12 mIoU), outperforming $I^2$-World
under CF-OOD.
COME exhibits lower consistency than both models across
all regimes, despite its scene-centric formulation.
Together with the qualitative observations in
Fig.~\ref{fig4}, these analyses reveal that action
faithfulness is not an all-or-nothing property.
Existing models respond to counterfactual actions, but
their forecasting accuracy and geometric consistency
deteriorate as the specified trajectories depart from
the training action distribution.
ReCast exhibits greater robustness to such shifts,
maintaining more accurate forecasts and higher geometric
consistency than the learned baselines under CF-OOD.

\subsection{Ablation Study}

Table~\ref{tab:alignment_ablation} evaluates the components of
ReCast while keeping the deformable attention operator fixed across variants.
Compared with the single spatiotemporal transformer in (a), decoupling the
historical encoder and future decoder in (b) improves mIoU by $+0.74$,
$+0.09$, and $+0.40$ on CF-ID, CF-Tail, and CF-OOD, respectively.
This result suggests a modest benefit of architectural
decoupling under counterfactual evaluation.
Introducing geometry-aligned temporal interactions in (c) improves
mIoU by $+19.25$, $+22.21$, and $+21.05$ across the three counterfactual
regimes, respectively.
The substantial gains under CF-Tail and CF-OOD suggest that
explicit geometric correspondence is beneficial
for counterfactual trajectories that depart from the
training action distribution.
Finally, incorporating causal future-to-future attention
provides additional, albeit modest, improvements across
all evaluation settings.

% NOTE: Archived checkpoints, corrected inference only; not action-index-corrected retraining.
\begin{table}[hbt!]
\centering
\setlength{\tabcolsep}{4pt}
% \renewcommand{\arraystretch}{1.08}
% \scriptsize
\resizebox{\linewidth}{!}{
\begin{tabular}{lcccc}
\toprule
Variant & Standard & CF-ID & CF-Tail & CF-OOD \\
\midrule

(a) Baseline
& 18.39 / 28.46
& 17.52 / 27.98
& 11.86 / 21.64
& 8.75 / 17.79
\\

(b) + Decoupled Arch.
& 18.12 / 27.97
& 18.26 / 27.98
& 11.95 / 21.65
& 9.15 / 18.01
\\

(c) + Geometry Alignment
& 39.33 / 48.53
& 37.51 / 46.56
& 34.16 / 43.50
& 30.20 / 39.69
\\

\textbf{(d) + Causal F2F (Ours)}
& \textbf{39.54} / \textbf{48.60}
& \textbf{37.79} / \textbf{46.57}
& \textbf{34.64} / \textbf{43.69}
& \textbf{31.13} / \textbf{40.33}
\\

\bottomrule
\end{tabular}
}
\caption{
Ablation of the architectural components of ReCast.
Results are averaged over six future horizons.
For the baseline, the block depth is doubled to match the parameter count.
}
\label{tab:alignment_ablation}
\vspace{-3.mm}
\end{table}

\begin{figure}[hbt!]
    \centering
    \includegraphics[width=\linewidth]{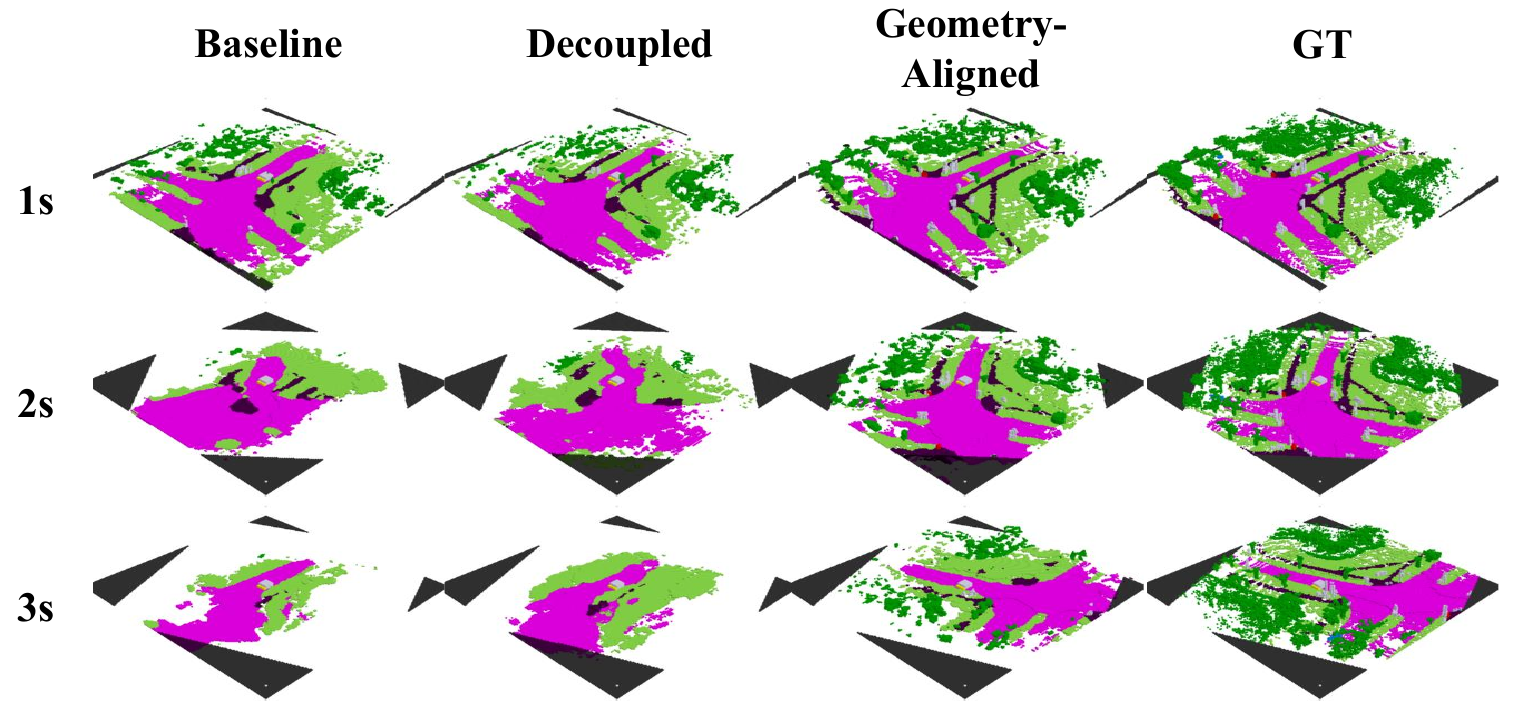}
    \caption{
    Qualitative ablation under a CF-OOD trajectory at 1, 2, and 3\,s.
    Baseline, Decoupled, and Geometry-Aligned correspond to
    variants (a), (b), and (c) in Table~\ref{tab:alignment_ablation}.
    }
    \label{fig6}
    \vspace{-3.mm}
\end{figure}

The qualitative comparison in Figure~\ref{fig6} further illustrates the effect
of the major architectural changes under a CF-OOD trajectory.
Compared with the baseline variant (a), architectural decoupling in (b)
improves adaptation to the specified ego motion.
However, at 2--3\,s, where the trajectory involves larger yaw changes, both
variants still exhibit limited adaptation to the specified action.
In contrast, explicitly grounding future prediction in geometric
correspondence in (c) yields future occupancy that follows the
counterfactual ego trajectory more consistently.
Together with the quantitative ablation, these results support the
effectiveness of ReCast's architectural design in improving action faithfulness.

\section{Conclusion}
\label{5_conclusion}

We investigated action faithfulness in occupancy world models,
revealing how scene--trajectory correlations can limit their
adaptation to counterfactual actions.
To systematically evaluate this limitation, we introduced a
counterfactual benchmark spanning in-distribution, tail, and
OOD ego trajectories.
We further proposed ReCast, which grounds future prediction
in action-induced geometric correspondence through a decoupled
encoder--decoder and geometry-aligned deformable attention.
Experiments on Occ3D-nuScenes demonstrate competitive standard
forecasting and improved counterfactual accuracy over learned
baselines, particularly under OOD actions.
Although our benchmark does not model surrounding agents'
reactions, it enables controlled evaluation of geometric
action faithfulness.
Future work will extend this framework toward interactive
scene dynamics while improving nominal forecasting without
compromising action faithfulness.

% \clearpage
% \section*{LLM Usage Statement}
% OpenAI ChatGPT was used to assist with research discussion,
% manuscript drafting, language editing, and interpretation
% of experimental analyses. OpenAI Codex assisted with code
% development, debugging, and evaluation-validation scripts.
% The authors reviewed the generated outputs and are responsible
% for the methodology, reported results, and conclusions.
% {
%     \small
%     \bibliographystyle{ieeenat_fullname}
%     \bibliography{main}
% }

% WARNING: do not forget to delete the supplementary pages from your submission 
% \clearpage
\setcounter{page}{1}
% \maketitlesupplementary
\appendix
\raggedbottom

% supplementary numbering: Table S1, Figure S1, Eq. (S1)
\setcounter{table}{0}
\setcounter{figure}{0}
\setcounter{equation}{0}
\renewcommand{\thetable}{S\arabic{table}}
\renewcommand{\thefigure}{S\arabic{figure}}
\renewcommand{\theequation}{S\arabic{equation}}

\section{Implementation Details}

We implement ReCast in PyTorch~2.0.1 with CUDA~11.8 and train it using distributed data parallelism on four RTX A6000 GPUs with BF16 mixed precision. Each sequence consists of four context frames and six future frames sampled at $0.5$\,s intervals.
Semantic occupancy is represented on a $200{\times}200{\times}16$ voxel grid
and compressed using the DOME~\cite{gu2024dome} occupancy VAE into a
$25{\times}25$ latent grid with 64 channels, scaled by a factor of
$0.18215$. During training, context latents are encoded from the ground-truth
occupancy, while future positions share a learnable latent token
initialized from the mean VAE latent of an empty occupancy grid with
Gaussian noise ($\sigma=10^{-4}$). The encoder and decoder use a feature dimension of 768 with 12 attention heads and an MLP ratio of 4. The encoder contains six blocks, while the decoder contains eight blocks. Spatial deformable attention uses eight sampling points, past-to-past and future-to-past attention use four sampling points for each reference point, and future-to-future attention uses eight sampling points. Learnable out-of-bound tokens are used for future-to-past retrieval. Training uses AdamW~\cite{loshchilov2017decoupled} with a learning rate of $2{\times}10^{-4}$, weight decay of $10^{-4}$, a global batch size of 32, and gradient clipping at $1.0$. The learning rate is linearly warmed up for 500 iterations from $10^{-6}$ and then cosine-decayed to $10^{-6}$. We train for 65,700 optimization iterations using a random seed of zero.

\section{Action-Faithfulness Evaluation Protocol}
\label{app:action_faithfulness}

\paragraph{Evaluation set construction.}
We construct the action-faithfulness evaluation set from all 150 scenes in the
Occ3D-nuScenes~\cite{tian2023occ3d} validation split.
For each scene, we use the first valid temporal window.
The recorded futures of these 150 windows serve as
nominal references for counterfactual construction.
The main-paper Standard evaluation separately uses
all valid validation windows. For each nominal
window, we generate three counterfactual trajectories for each of CF-ID,
CF-Tail, and CF-OOD, yielding 150 nominal and 1,350 counterfactual samples.
Each future trajectory is represented by six ego-frame $\mathrm{SE}(2)$
relative motions, whose cumulative poses are flattened into an 18-dimensional
feature vector for density estimation. All density statistics and partition
thresholds are computed exclusively from the training split.

\paragraph{Trajectory-density partitioning.}
We model the training action distribution using a $K$-nearest-neighbor density
score. From the 700 training scenes, we extract 11,122 trajectories
using sliding 10-frame windows with a temporal stride of two. Each trajectory
dimension is standardized using the training-set mean and standard deviation.
The density score is the mean Euclidean distance to the $K_{\mathrm{nn}}=30$ nearest training
trajectories in the normalized feature space, where larger values indicate
lower training density. Self-neighbors are excluded only when estimating the
training-score distribution.
The strict partitions are defined using the training-score
quantiles in Table~\ref{tab:cf_partition}. CF-ID contains trajectories at or below $\tau_{90}$, CF-Tail contains trajectories between $\tau_{95}$ and $\tau_{99}$,
and CF-OOD contains trajectories above $\tau_{99}$. The interval
$(\tau_{90},\tau_{95}]$ is excluded from the strict definition and used only
during relaxed Tail filling. Nominal validation windows are not density
filtered: 131 of 150 fall below $\tau_{90}$, while 19 lie in the
training-distribution tail.
\begin{table}[t]
\centering
\setlength{\tabcolsep}{10pt}
\resizebox{\linewidth}{!}{
\begin{tabular}{lcc}
\toprule
Partition & Nominal density range & Retained \\
\midrule
CF-ID
& $s(\mathbf{a}) \leq \tau_{90}$
& 450 \\
CF-Tail
& $\tau_{95}<s(\mathbf{a})\leq\tau_{99}$
& 450 \\
CF-OOD
& $s(\mathbf{a})>\tau_{99}$
& 450 \\
\midrule
\multicolumn{1}{l}{Thresholds}
& \multicolumn{2}{l}{
$\tau_{90}=1.254,\;
 \tau_{95}=1.607,\;
 \tau_{99}=2.457$} \\
\bottomrule
\end{tabular}}
\caption{Nominal density partitions and retained counterfactual samples.
Seven CF-Tail samples are retained through relaxed density-based completion.}
\label{tab:cf_partition}
\end{table}
The ranges in Table~\ref{tab:cf_partition} define the nominal
density partitions. Seven of the 450 retained CF-Tail
trajectories fall outside the strict interval due to
relaxed proposal completion. All 1,350 counterfactual
samples are retained for evaluation.

We examine the sensitivity of partition assignments to
the neighborhood size and quantile thresholds.
Table~\ref{tab:cf_knn} reports agreement with the retained benchmark
assignments when the neighborhood size changes, with thresholds recomputed
for each $K_{\mathrm{nn}}$.
Nearby choices give similar assignments; this agreement measures partition
stability relative to the retained benchmark rather than the accuracy of the
density estimator.
\begin{table}[t]
\centering
\setlength{\tabcolsep}{10pt}
\resizebox{\linewidth}{!}{
\begin{tabular}{ccccc}
\toprule
$K_{\mathrm{nn}}$
& $\tau_{90}$
& $\tau_{95}$
& $\tau_{99}$
& Agreement (\%) \\
\midrule
5   & 0.824 & 1.065 & 1.679 & 77.9 \\
10  & 0.954 & 1.235 & 1.923 & 80.9 \\
20  & 1.130 & 1.447 & 2.240 & 94.6 \\
30  & 1.254 & 1.607 & 2.457 & 99.5 \\
50  & 1.440 & 1.849 & 2.833 & 97.6 \\
100 & 1.770 & 2.296 & 3.395 & 94.7 \\
\bottomrule
\end{tabular}}
\caption{Sensitivity to the number of nearest neighbors.
Thresholds are recomputed for each $K_{\mathrm{nn}}$.}
\label{tab:cf_knn}
\end{table}
The assignments remain stable under small changes to quantile boundaries,
whereas larger shifts alter the partition semantics, as shown in
Table~\ref{tab:cf_quantile}.

\begin{table}[t]
\centering
\setlength{\tabcolsep}{3pt}
\resizebox{\linewidth}{!}{
\begin{tabular}{lcc}
\toprule
Quantiles & Corresponding thresholds & Agreement (\%) \\
\midrule
90 / 95 / 99
& 1.254 / 1.607 / 2.457
& 99.5 \\
92 / 96 / 99
& 1.392 / 1.691 / 2.457
& 97.3 \\
90 / 97 / 99.5
& 1.254 / 1.822 / 3.071
& 91.0 \\
85 / 92 / 97
& 0.946 / 1.392 / 1.822
& 62.1 \\
\bottomrule
\end{tabular}}
\caption{Density-quantile sensitivity with $K_{\mathrm{nn}}=30$.}
\label{tab:cf_quantile}
\end{table}

\paragraph{Counterfactual action generation.}
For each nominal trajectory, we generate counterfactual candidates using
speed, curvature, and compound perturbations, smoothly introduced over the
six-step horizon according to
\begin{equation}
e_k
=
\frac{1}{2}
-
\frac{1}{2}
\cos\left(\pi\frac{k-1}{5}\right),
\qquad
k\in\{1,\ldots,6\}.
\end{equation}
The perturbation magnitude at step $k$ is scaled by $e_k$, increasing from zero
at the first future step to its full value at the last. Speed perturbations
modify planar translation, curvature perturbations modify yaw, and compound
perturbations modify both. CF-ID uses mild perturbations throughout. CF-Tail
uses mild ranges and explores stronger perturbations in later
proposal rounds, whereas CF-OOD uses the stronger ranges throughout. For near-stationary trajectories with a mean per-step
displacement below $0.15$\,m, we add a smooth
translational displacement to avoid degenerate proposals.
Candidates that are nearly identical to the nominal trajectory or duplicate
another candidate from the same scene are discarded before density-based
partition assignment.

Table~\ref{tab:cf_action_magnitude} summarizes the resulting action
distributions. Density score and yaw perturbation increase from CF-ID to
CF-Tail and CF-OOD, while translation perturbations are more variable,
particularly for CF-ID samples obtained through trajectory blending.
\begin{table*}[t]
\centering
% \small
\setlength{\tabcolsep}{5pt}
\begin{tabular}{lccccccccc}
\toprule
& \multicolumn{3}{c}{CF-ID}
& \multicolumn{3}{c}{CF-Tail}
& \multicolumn{3}{c}{CF-OOD} \\
\cmidrule(lr){2-4}
\cmidrule(lr){5-7}
\cmidrule(lr){8-10}
Metric
& Mean & Med. & P90
& Mean & Med. & P90
& Mean & Med. & P90 \\
\midrule
Density score
& 0.68 & 0.66 & 1.11
& 2.07 & 2.10 & 2.40
& 6.46 & 5.58 & 9.91 \\

$|\Delta x|$ (m/step)
& 0.05 & 0.02 & 0.18
& 0.07 & 0.02 & 0.40
& 0.15 & 0.03 & 0.40 \\

$|\Delta y|$ (m/step)
& 1.20 & 1.18 & 2.48
& 1.09 & 0.14 & 3.10
& 2.05 & 0.16 & 6.31 \\

$|\Delta v|$ (m/step)
& 1.22 & 1.18 & 2.48
& 1.14 & 0.41 & 3.10
& 2.16 & 1.01 & 6.32 \\

$|\Delta\psi|$ (rad/step)
& 0.01 & 0.00 & 0.04
& 0.10 & 0.12 & 0.15
& 0.20 & 0.25 & 0.35 \\

$\sum_k |\Delta v_k|$ (m)
& 7.3 & 7.1 & 14.9
& 6.8 & 2.5 & 18.6
& 13.0 & 6.0 & 37.9 \\

$\sum_k |\Delta\psi_k|$ (rad)
& 0.07 & 0.00 & 0.27
& 0.59 & 0.73 & 0.90
& 1.20 & 1.50 & 2.09 \\

Median stepwise speed ratio
& 1.02 & 1.00 & 1.80
& 1.25 & 1.00 & 2.70
& 2.02 & 1.80 & 3.00 \\
\bottomrule
\end{tabular}
\caption{Counterfactual action magnitude across density regimes.
Mean, median (Med.), and 90th percentile (P90) are
reported for each metric. The speed-ratio row aggregates
per-trajectory median stepwise ratios.}
\label{tab:cf_action_magnitude}
\end{table*}

For the speed-ratio statistic, we first compute the median stepwise ratio for
each trajectory using a nominal-speed floor of $0.15$\,m/s and then aggregate
the sample-level values. The accumulated yaw term
$\sum_k|\Delta\psi_k|$ denotes the summed absolute yaw-action difference rather
than the final heading difference.

\paragraph{Counterfactual occupancy generation.}
For each accepted trajectory, we construct a counterfactual occupancy target by
re-expressing the recorded nominal future occupancy in the substituted ego
frame. This preserves the nominal evolution of surrounding agents and changes
only the reference frame induced by the specified ego trajectory.
Let $\mathbf{T}^{\mathrm{cf}\leftarrow\mathrm{nom}}_{t+\tau}$ map
coordinates from the nominal future ego frame into the counterfactual future
ego frame, as in Eq.~(23) of the main paper. The two cumulative ego poses at
horizon $\tau$ determine this relative $\mathrm{SE}(2)$ transformation.
For inverse-map voxel resampling, the source coordinate corresponding to a
counterfactual target position $\mathbf{p}^{\mathrm{cf}}$ is
\begin{equation}
\label{eq:cf_inverse_sampling}
\begin{aligned}
\mathbf{T}^{\mathrm{nom}\leftarrow\mathrm{cf}}_{t+\tau}
&=\bigl(\mathbf{T}^{\mathrm{cf}\leftarrow\mathrm{nom}}_{t+\tau}\bigr)^{-1},\\
\mathbf{p}^{\mathrm{nom}}
&=\mathcal{G}\bigl(\mathbf{p}^{\mathrm{cf}};
  \mathbf{T}^{\mathrm{nom}\leftarrow\mathrm{cf}}_{t+\tau}\bigr),\\
\mathbf{O}^{\mathrm{cf}}_{t+\tau}(\mathbf{p}^{\mathrm{cf}})
&=\mathbf{O}^{\mathrm{nom}}_{t+\tau}(\mathbf{p}^{\mathrm{nom}}).
\end{aligned}
\end{equation}
We use nearest-neighbor resampling for semantic occupancy labels.
Coordinates whose source positions lie outside the nominal occupancy grid
are excluded by the counterfactual validity mask; the mask is transformed
consistently with the target. The corrected inverse mapping, target grids,
and validity masks are checked for all 1,350 cases. Round-trip checks are
reported separately below and do not redefine the evaluation population.

\paragraph{Scope of counterfactual evaluation.}
Our benchmark isolates geometric action faithfulness rather than
simulating the full causal consequences of alternative ego actions.
Counterfactual targets preserve the recorded scene evolution while
re-expressing future occupancy under substituted ego trajectories.
The protocol therefore evaluates whether predictions respect
specification-induced geometric correspondence, but does not measure
how other agents would react to alternative ego behavior.
Moreover, passing the benchmark does not establish the physical
feasibility or safety of a counterfactual trajectory.

\paragraph{Multi-stage construction and quality control.}
Counterfactual trajectories are generated through a main pass,
a kinematic retry, and a trajectory-blending fallback.
As noted above, seven CF-Tail samples lie below the strict
$\tau_{95}$ threshold (density scores: 1.278--1.583) under relaxed
density-based proposal completion: three curvature, one speed, and
three training-trajectory blending proposals.
All 1,350 counterfactual samples are retained for evaluation,
regardless of post-hoc QC outcomes.

\paragraph{Geometric verification.}
We check the counterfactual occupancy targets and validity masks
for all 1,350 samples against the independently corrected
geometric evaluation artifacts.
Using the corrected transformations and transported common support,
we compute mean valid coverage $V$ across the six horizons.
Static-column round-trip exactness $E$ measures the fraction of
GT-static columns whose labels match at every height, while
static-region mIoU $M$ is computed over GT-static voxels
within common support, aggregating six horizons.
Both QC configurations require the native screening predicate $P$,
which holds for all 1,350 evaluated samples.
Strict geometric QC requires
$P\land(V\geq0.55)\land(E\geq0.96)\land(M\geq0.90)$,
and 1,231 samples satisfy these criteria.
Of the 119 strict failures, 114 are attributed to exactness
and five to mIoU under the sequential $P\to V\to E\to M$ check.
Uniformly relaxed QC retains the coverage threshold and requires
$P\land(V\geq0.55)\land(E\geq0.93)\land(M\geq0.87)$,
yielding 1,342 passing samples (445, 449, and 448 for
CF-ID, CF-Tail, and CF-OOD, respectively).
These geometric QC results are diagnostic and do not
alter the evaluation population or validity masks.

\paragraph{Kinematic diagnostics.}
We additionally assess trajectory kinematics using
scalar- and vector-based SI-unit checks with a fixed
time interval of $\Delta t=0.5$\,s.
The original proposal screening used per-step displacement
as a speed proxy, rather than consistently converting
displacement into velocity.
We therefore distinguish the original screening from
the independent post-hoc kinematic diagnostics.
The scalar and vector checks pass 1,243 and 1,141
samples, respectively; 841 and 783 also pass the corresponding
historical-motion continuity checks.
Because these diagnostics assume a fixed time interval
rather than actual timestamps, they do not certify
physical feasibility for every trajectory.
Table~\ref{tab:cf_qc} summarizes the verification results.

\begin{table}[t]
\centering
\setlength{\tabcolsep}{3pt}
\resizebox{\linewidth}{!}{
\begin{tabular}{lrrrr}
\toprule
Diagnostic & CF-ID & CF-Tail & CF-OOD & Total \\
\midrule
Evaluation population & 450 & 450 & 450 & 1,350 \\
Strict density & 450 & 443 & 450 & 1,343 \\
Corrected strict geometry & 436 & 399 & 396 & 1,231 \\
Uniform relaxed geometry & 445 & 449 & 448 & 1,342 \\
SI scalar ($\Delta t=0.5$\,s) & 445 & 429 & 369 & 1,243 \\
SI vector ($\Delta t=0.5$\,s) & 445 & 429 & 267 & 1,141 \\
Scalar + history join & 268 & 316 & 257 & 841 \\
Vector + history join & 268 & 316 & 199 & 783 \\
\bottomrule
\end{tabular}
}
\caption{
Counterfactual benchmark quality-control summary.
All 1,350 samples are evaluated, including those failing
post-hoc geometric or kinematic diagnostics.
Kinematic checks assume a fixed 0.5\,s interval rather
than actual timestamps.
}
\label{tab:cf_qc}
\end{table}

\paragraph{Robustness across QC-conditioned subsets.}
We re-evaluate the same archived model predictions after restricting the
counterfactual evaluation population to samples passing corrected strict
geometric QC (G), its intersection with the fixed-interval SI vector check
(G+SI), and the further intersection with historical-motion continuity
(G+SI+H). The resulting populations contain 1,231, 1,051, and 714
trajectories, respectively, compared with 1,350 in the full benchmark.
All models use identical subset membership, corrected targets, and validity
masks, with the original six-horizon metric aggregation. Table~\ref{tab:qc_robustness}
shows that ReCast maintains a positive Full-region mIoU difference over
$I^2$-World~\cite{liao2025i2} in each regime and subset. In contrast, rigid propagation
remains stronger under CF-OOD, and QC filtering does not establish the
superiority of ReCast over this training-free reference. Because filtering
also changes the action and scene distributions, these results are
sensitivity checks rather than estimates of a causal QC effect.

\begin{table*}[t]
\centering
% \small
\setlength{\tabcolsep}{5pt}
\begin{tabular}{llr rrrr rr}
\toprule
Subset & Regime & Samples & ReCast & $I^2$-World & Rigid & COME
& $\Delta$ vs. $I^2$ & $\Delta$ vs. Rigid \\
\midrule
All & CF-ID   & 450 & 37.79 & 34.79 & 36.24 & 31.83 & +3.01 & +1.55 \\
All & CF-Tail & 450 & 34.64 & 32.20 & 35.58 & 21.29 & +2.45 & -0.94 \\
All & CF-OOD  & 450 & 31.13 & 23.14 & 35.85 & 15.97 & +7.99 & -4.72 \\
\midrule
G & CF-ID   & 436 & 37.97 & 34.95 & 36.51 & 32.03 & +3.02 & +1.45 \\
G & CF-Tail & 399 & 34.94 & 32.63 & 36.15 & 21.46 & +2.31 & -1.20 \\
G & CF-OOD  & 396 & 31.61 & 23.59 & 36.80 & 16.28 & +8.02 & -5.19 \\
\midrule
G+SI & CF-ID   & 431 & 38.03 & 34.99 & 36.58 & 32.11 & +3.03 & +1.45 \\
G+SI & CF-Tail & 378 & 35.15 & 32.81 & 36.40 & 21.54 & +2.34 & -1.25 \\
G+SI & CF-OOD  & 242 & 33.22 & 25.97 & 40.74 & 18.47 & +7.25 & -7.52 \\
\midrule
G+SI+H & CF-ID   & 260 & 40.64 & 37.26 & 41.28 & 34.41 & +3.38 & -0.64 \\
G+SI+H & CF-Tail & 274 & 35.86 & 33.48 & 38.67 & 23.00 & +2.38 & -2.81 \\
G+SI+H & CF-OOD  & 180 & 34.94 & 26.93 & 45.00 & 19.87 & +8.01 & -10.06 \\
\bottomrule
\end{tabular}
\caption{Full-region semantic mIoU (\%) under counterfactual QC filtering,
averaged over six horizons. $\Delta$ denotes ReCast minus the indicated
comparator (percentage points). G: corrected strict geometric QC;
SI: fixed-$0.5$\,s vector-based kinematic check; H: historical-motion
continuity check. All four methods use the same retained samples in
any given row. These are diagnostic subsets, not certified feasible
trajectories. Rigid uses corrected nearest-neighbor propagation.}
\label{tab:qc_robustness}
\end{table*}

\paragraph{Descriptive action-severity and coverage checks.}
Within CF-OOD, ReCast exceeds $I^2$-World by 5.40, 8.82, and 10.15
Full-region mIoU points across increasing density-score tertiles.
Across increasing corrected-valid-coverage tertiles, the corresponding
paired gaps are 8.09, 8.25, and 7.53 points. These comparisons use
identical samples and masks within each stratum, but the strata differ
in scene and action composition and have no separate significance tests.

\section{Evaluation Metrics and Statistical Analysis}
\label{app:evaluation_metrics}

\paragraph{Evaluation population and aggregation.}
The main-paper Standard comparison reports results at
1, 2, and 3\,s, while the regional analysis uses
six-horizon averages. The counterfactual benchmark
evaluates 1,350 trajectories derived from 150 fixed
scene anchors over all six future horizons.
We report counterfactual mIoU and binary occupancy IoU as the
arithmetic mean over these six horizons, after accumulating the voxel-level
intersection and union counts across the evaluation cases at each horizon.
Only target voxels valid under the corrected counterfactual transformation
contribute to each score.
All 17 semantic classes have pooled ground-truth support at every evaluated
horizon. Consequently, the legacy and union-supported Full mIoU conventions
produce identical scores and model rankings for the reported populations.

\paragraph{Region definitions.}
\emph{Full} covers all valid target voxels. \emph{Change} restricts this
mask to voxel indices where the recorded nominal and counterfactual future
semantic labels differ. This region includes changes induced by the
coordinate transformation and voxel rasterization; it is not a direct
measure of reactive changes in other agents.
\emph{Dynamic} semantic mIoU averages the fixed set of eight dynamic
classes. Class-wise counts are pooled across all evaluation cases
separately at each horizon. Following the legacy evaluator, a class
with zero ground-truth count is assigned IoU~1, even when false-positive
predictions are present, and remains included in the class average.
The resulting scores are averaged over the six future horizons.
Binary occupancy IoU is computed on the ground-truth dynamic ROI.
\emph{Unseen} is the valid region outside the union of the geometric
footprints of all four historical occupancy grids when mapped into each
future ego frame, without patch expansion. It therefore measures geometric
out-of-support prediction, not sensor occlusion. Semantic mIoU in this
region uses union-supported classes; this differs from the legacy convention
used for the main Full-region benchmark scores.

\paragraph{Sensitivity to the dynamic-class convention.}
We recompute dynamic semantic mIoU using a union-supported class set,
which penalizes predicted-only classes rather than assigning them IoU~1.
For each horizon in the Standard and three counterfactual populations,
all eight dynamic classes have nonzero ground-truth support after pooling
counts across evaluation cases. Consequently, the union-supported,
common-GT-supported, and legacy dynamic mIoU values are identical for
all compared models. This leaves the model ordering unchanged:
$I^2$-World exceeds ReCast by 0.60 points under Standard forecasting,
whereas ReCast exceeds $I^2$-World by 1.64, 0.74, and 6.11 points under
CF-ID, CF-Tail, and CF-OOD, respectively. Rigid propagation retains a
higher CF-OOD dynamic mIoU than ReCast (29.26 vs. 27.75).
These checks reuse archived predictions and do not modify the main-paper
metric definitions or results.

\paragraph{Regional statistical uncertainty.}
We compare ReCast and $I^2$-World on matched scenes using paired
scene-bootstrap confidence intervals (CIs). Table~\ref{tab:unseen_ci}
reports the differences in Unseen scores; CIs crossing zero do not establish
a positive difference, and no equivalence margin is assumed.
Unseen semantic improvements are supported by the interval only under
CF-OOD, while binary occupancy IoU shows positive differences in all three
regimes.

\begin{table}[t]
\centering
% \small
% \setlength{\tabcolsep}{1pt}
% \renewcommand{\arraystretch}{1.12}
\resizebox{\linewidth}{!}{
\begin{tabular*}{\linewidth}{@{\extracolsep{\fill}}lcc@{}}
\toprule
Regime & Semantic mIoU & Occupancy IoU \\
\midrule
CF-ID   & $+0.71\;[-0.41,\,2.08]$ & $+3.72\;[2.63,\,4.76]$ \\
CF-Tail & $+0.37\;[-0.39,\,1.49]$ & $+2.90\;[1.87,\,3.90]$ \\
CF-OOD  & $+1.19\;[0.61,\,1.94]$ & $+3.87\;[2.93,\,4.83]$ \\
\bottomrule
\end{tabular*}}
\caption{ReCast minus $I^2$-World on Unseen regions.
Entries report differences in percentage points with
95\% paired scene-bootstrap confidence intervals.}
\label{tab:unseen_ci}
\end{table}

\paragraph{Full-region statistical uncertainty.}
We also assess the primary Full-region comparisons with 5,000 paired
scene-cluster bootstrap replicates (PCG64 seed 20261124), pooling voxel
counts within each horizon before averaging six horizon scores. Each
bootstrap draw resamples scenes jointly for the compared models.
As shown in Table~\ref{tab:full_ci}, ReCast improves over $I^2$-World
in all three regimes. The corresponding Full-region occupancy IoU differences against
$I^2$-World are +5.76, +5.23, and +11.18 points for CF-ID,
CF-Tail, and CF-OOD, respectively, with positive paired 95\% CIs.
The comparison against rigid propagation is
more limited: CF-ID is borderline, CF-Tail is inconclusive, and rigid
propagation is better under CF-OOD. The intervals condition on fixed
model checkpoints and evaluation samples; they are not adjusted for
multiple comparisons or training-seed variability.

\begin{table}[t]
\centering
\small
\setlength{\tabcolsep}{7pt}
\resizebox{\linewidth}{!}{
\begin{tabular}{lcc}
\toprule
Regime & vs. $I^2$-World & vs. Rigid \\
\midrule
CF-ID   & $+3.01\,[2.22,3.71]$ & $+1.55\,[0.003,2.85]$ \\
CF-Tail & $+2.45\,[1.78,3.03]$ & $-0.94\,[-2.75,0.57]$ \\
CF-OOD  & $+7.99\,[7.13,8.77]$ & $-4.72\,[-6.60,-3.05]$ \\
\bottomrule
\end{tabular}}
\caption{ReCast minus comparator in Full-region semantic mIoU
(percentage points), with 95\% paired scene-bootstrap CIs.
The near-zero CF-ID lower bound against rigid propagation is
sensitive to Monte Carlo resampling.}
\label{tab:full_ci}
\end{table}

\section{Action Intervention and Geometric References}
\label{app:intervention}

\paragraph{Matched and swapped actions.}
For each scene and density regime, three distinct counterfactual ego
actions are available. We evaluate each action against its corresponding
target (matched), then replace that action with either of the other two
actions from the same regime while keeping the observed history, target,
and validity mask fixed (swapped). This yields six ordered off-diagonal
action pairs per scene and regime, or
$150\times3\times6=2{,}700$ ordered swaps. Matched and swapped accuracy
are evaluated against the same ground truth for each target action.

\paragraph{Cross-action geometric consistency.}
For each pair, we also predict separately under the two actions, warp the
predictions into a common ego-coordinate frame using their relative
$\mathrm{SE}(2)$ pose, and measure semantic and binary occupancy agreement
on valid common support. Unlike matched/swapped accuracy, this consistency
score compares the two predictions with each other rather than with a
future occupancy target. Results are averaged over the six horizons;
Figure~5 of the main paper reports 95\% scene-bootstrap CIs.

\paragraph{Training-free rigid propagation.}
The rigid reference transforms the last observed occupancy into each
specified future ego frame using geometric resampling, without any learned
future scene evolution. Its predictions are evaluated under the same
counterfactual target and validity masks as the learned models.
Because rigid propagation can only transport historical occupied content,
it cannot produce occupied predictions beyond its historical geometric
source support. This explains its zero occupied predictions in the strict
Unseen region, even when it achieves strong Full- or Change-region scores.

\paragraph{Horizon-dependent performance.}
We further examine Full-region mIoU across the six prediction horizons
using the original evaluation populations and aggregation protocol.
Under CF-OOD, ReCast's advantage over $I^2$-World increases from
0.80 points at 0.5\,s to 12.79 points at 3\,s, while its difference
from rigid propagation changes from +0.02 to -11.09 points.
Similar longer-horizon gains over $I^2$-World are observed under CF-ID
and CF-Tail, whereas ReCast remains below $I^2$-World at all standard
forecasting horizons. These are descriptive fixed-checkpoint comparisons;
no significance test of the change in gap across horizons is implied.
The results highlight comparator-dependent temporal behavior rather than
a universal advantage over geometric propagation.

% \section{Additional Qualitative Results}

% \begin{figure*}[t]
%     \centering
%     \includegraphics[width=\linewidth]{Figures/fig7.pdf}
%     \caption{Additional qualitative comparison between $I^2$-World and ReCast under
% standard and counterfactual trajectories at
% 1\,s and 3\,s. Black regions denote voxels outside the valid occupancy range
% after counterfactual transformation.}
%     \label{fig7}
% \end{figure*}
% Figure~\ref{fig7} presents additional qualitative comparisons between
% $I^2$-World and ReCast on five representative scenes under
% the standard, CF-ID, CF-Tail, and CF-OOD settings.
% Under the standard and CF-ID settings, both methods generally produce
% reasonable future occupancies. As the counterfactual trajectory becomes less
% likely, however, their behaviors diverge. In the first three examples,
% $I^2$-World degrades substantially under CF-OOD, whereas ReCast preserves
% coherent road layouts and scene geometry while remaining consistent with the
% specified counterfactual trajectory.
% The last two examples highlight complementary CF-OOD cases. The fourth
% contains only a mild lateral translation with little rotational change, yet
% $I^2$-World still produces substantial geometric degradation, whereas ReCast
% continues to generate plausible occupancy predictions. The fifth illustrates
% an extreme compound perturbation. Although ReCast exhibits some translation
% misalignment under this severe action shift, it still preserves the overall
% road topology and scene structure, while $I^2$-World largely fails to maintain
% a coherent prediction.

\section{Geometry-Aligned Temporal Attention Algorithms}
\label{app:algorithms}

Algorithms~\ref{alg:p2p}--\ref{alg:f2f} summarize the three geometry-aligned
temporal attention modules used in ReCast. We follow the notation of the main
manuscript, where historical frame indices are
$i,j\in\{t-3,\ldots,t\}$, future horizons are
$\tau\in\{1,\ldots,T_f\}$, and
$\mathcal{G}(\mathbf{p};\mathbf{T}_{b\leftarrow a})$
maps a BEV position from frame $a$ to frame $b$.
Deformable attention predicts sampling offsets and attention weights according
to Eq.~(4) in the main manuscript.

\begin{algorithm}[hbt!]
\footnotesize
\caption{Past-to-Past Geometry-Aligned Attention}
\label{alg:p2p}
\begin{algorithmic}[1]
\REQUIRE Spatially refined historical features
$\{\mathbf{H}^{\mathrm{sp},\ell}_{i}\}_{i=t-3}^{t}$;
relative transformations
$\{\mathbf{T}_{j\leftarrow i}\}$;
BEV positions $\mathbf{p}$
\ENSURE Temporally refined features
$\{\mathbf{H}^{\mathrm{tmp},\ell}_{i}\}_{i=t-3}^{t}$

\STATE Construct the historical feature set
\[
\mathbf{H}^{\ell}_{\mathrm{hist}}
=
\left\{
\mathbf{H}^{\mathrm{sp},\ell}_{j}
\right\}_{j=t-3}^{t}.
\]

\FOR{$i=t-3,\ldots,t$}
    \FOR{each BEV position $\mathbf{p}$}
        \FOR{$j=t-3,\ldots,t$}
            \STATE Compute the aligned reference
            \[
            \overline{\mathbf{p}}^{\mathrm{hist}}_{j\leftarrow i}
            (\mathbf{p})
            =
            \mathcal{G}
            \left(
            \mathbf{p};
            \mathbf{T}_{j\leftarrow i}
            \right).
            \]

            \STATE Construct its deformable sampling set
            \[
            \mathcal{P}^{\mathrm{hist}}_{j\leftarrow i}(\mathbf{p})
            =
            \left\{
            \overline{\mathbf{p}}^{\mathrm{hist}}_{j\leftarrow i}
            (\mathbf{p})
            +
            \Delta\mathbf{p}^{\mathrm{hist},(k)}_{j\leftarrow i}
            (\mathbf{p})
            \right\}_{k=1}^{K_{\mathrm{hist}}}.
            \]
        \ENDFOR

        \STATE Collect sampling sets from all historical frames
        \[
        \mathcal{P}^{\mathrm{hist}}_{i}
        =
        \left\{
        \mathcal{P}^{\mathrm{hist}}_{j\leftarrow i}
        \right\}_{j=t-3}^{t}.
        \]
    \ENDFOR

    \STATE Apply geometry-aligned temporal attention
    \[
    \mathbf{H}^{\mathrm{tmp},\ell}_{i}
    =
    \mathbf{H}^{\mathrm{sp},\ell}_{i}
    +
    \mathcal{D}_{\mathrm{hist}}
    \left(
    \mathbf{H}^{\mathrm{sp},\ell}_{i},
    \mathbf{H}^{\ell}_{\mathrm{hist}},
    \mathbf{H}^{\ell}_{\mathrm{hist}};
    \mathcal{P}^{\mathrm{hist}}_{i}
    \right).
    \]
\ENDFOR

\RETURN
$\{\mathbf{H}^{\mathrm{tmp},\ell}_{i}\}_{i=t-3}^{t}$
\end{algorithmic}
\end{algorithm}

\begin{algorithm}[hbt!]
\footnotesize
\caption{Future-to-Past Geometry-Aligned Attention}
\label{alg:f2p}
\begin{algorithmic}[1]
\REQUIRE Future query features
$\{\mathbf{Q}^{\ell}_{\tau}\}_{\tau=1}^{T_f}$;
context memory $\mathbf{H}^{\mathrm{ctx}}$;
relative transformations
$\{\mathbf{T}_{i\leftarrow t+\tau}\}$;
invalid-context token $\mathbf{e}_{\mathrm{inv}}$
\ENSURE Context-enhanced future features
$\{\widetilde{\mathbf{Q}}^{\ell}_{\tau}\}_{\tau=1}^{T_f}$

\FOR{$\tau=1,\ldots,T_f$}
    \FOR{each BEV position $\mathbf{p}$}
        \FOR{$i=t-3,\ldots,t$}
            \STATE Map the future position into historical frame $i$
            \[
            \overline{\mathbf{p}}^{\mathrm{ctx}}_{i\leftarrow\tau}
            (\mathbf{p})
            =
            \mathcal{G}
            \left(
            \mathbf{p};
            \mathbf{T}_{i\leftarrow t+\tau}
            \right).
            \]

            \STATE Construct the corresponding sampling set
            \[
            \mathcal{P}^{\mathrm{ctx}}_{i\leftarrow\tau}(\mathbf{p})
            =
            \left\{
            \overline{\mathbf{p}}^{\mathrm{ctx}}_{i\leftarrow\tau}
            (\mathbf{p})
            +
            \Delta\mathbf{p}^{\mathrm{ctx},(k)}_{i\leftarrow\tau}
            (\mathbf{p})
            \right\}_{k=1}^{K_{\mathrm{ctx}}}.
            \]
        \ENDFOR

        \STATE Collect sampling sets from all historical frames
        \[
        \mathcal{P}^{\mathrm{ctx}}_{\tau}
        =
        \left\{
        \mathcal{P}^{\mathrm{ctx}}_{i\leftarrow\tau}
        \right\}_{i=t-3}^{t}.
        \]
    \ENDFOR

    \STATE Retrieve historical evidence
    \[
    \mathbf{A}^{\ell}_{\tau}
    =
    \mathcal{D}_{\mathrm{ctx}}
    \left(
    \mathbf{Q}^{\ell}_{\tau},
    \mathbf{H}^{\mathrm{ctx}},
    \mathbf{H}^{\mathrm{ctx}};
    \mathcal{P}^{\mathrm{ctx}}_{\tau}
    \right).
    \]

    \STATE Identify positions with no valid historical sample

    \STATE Replace their attention outputs with
    $\mathbf{e}_{\mathrm{inv}}$

    \STATE Apply the residual update
    \[
    \widetilde{\mathbf{Q}}^{\ell}_{\tau}
    =
    \mathbf{Q}^{\ell}_{\tau}
    +
    \mathbf{A}^{\ell}_{\tau}.
    \]
\ENDFOR

\RETURN
$\{\widetilde{\mathbf{Q}}^{\ell}_{\tau}\}_{\tau=1}^{T_f}$
\end{algorithmic}
\end{algorithm}

\begin{algorithm}[hbt!]
\footnotesize
\caption{Causal Future-to-Future Geometry-Aligned Temporal Attention}
\label{alg:f2f}
\begin{algorithmic}[1]
\REQUIRE Spatially refined future features
$\{\mathbf{Q}^{\mathrm{sp},\ell}_{\tau}\}_{\tau=1}^{T_f}$;
relative transformations
$\{\mathbf{T}_{t+\tau-1\leftarrow t+\tau}\}$
\ENSURE Temporally refined future features
$\{\mathbf{Q}^{\mathrm{tmp},\ell}_{\tau}\}_{\tau=1}^{T_f}$

\STATE Preserve the first future horizon
\[
\mathbf{Q}^{\mathrm{tmp},\ell}_{1}
=
\mathbf{Q}^{\mathrm{sp},\ell}_{1}.
\]

\FOR{$\tau=2,\ldots,T_f$}
    \FOR{each BEV position $\mathbf{p}$}
        \STATE Map the current position into the preceding future frame
        \[
        \overline{\mathbf{p}}^{\mathrm{fut}}_{\tau-1\leftarrow\tau}
        (\mathbf{p})
        =
        \mathcal{G}
        \left(
        \mathbf{p};
        \mathbf{T}_{t+\tau-1\leftarrow t+\tau}
        \right).
        \]

        \STATE Construct the deformable sampling set
        \[
        \mathcal{P}^{\mathrm{fut}}_{\tau-1\leftarrow\tau}
        (\mathbf{p})
        =
        \left\{
        \overline{\mathbf{p}}^{\mathrm{fut}}_{\tau-1\leftarrow\tau}
        (\mathbf{p})
        +
        \Delta\mathbf{p}^{\mathrm{fut},(k)}_{\tau-1\leftarrow\tau}
        (\mathbf{p})
        \right\}_{k=1}^{K_{\mathrm{fut}}}.
        \]
    \ENDFOR

    \STATE Retrieve features from the preceding future frame
    \[
    \mathbf{Q}^{\mathrm{tmp},\ell}_{\tau}
    =
    \mathbf{Q}^{\mathrm{sp},\ell}_{\tau}
    +
    \mathcal{D}_{\mathrm{fut}}
    \left(
    \mathbf{Q}^{\mathrm{sp},\ell}_{\tau},
    \mathbf{Q}^{\mathrm{sp},\ell}_{\tau-1},
    \mathbf{Q}^{\mathrm{sp},\ell}_{\tau-1};
    \mathcal{P}^{\mathrm{fut}}_{\tau-1\leftarrow\tau}
    \right).
    \]
\ENDFOR

\RETURN
$\{\mathbf{Q}^{\mathrm{tmp},\ell}_{\tau}\}_{\tau=1}^{T_f}$
\end{algorithmic}
\end{algorithm}

\clearpage
{
    \small
    \bibliographystyle{ieeenat_fullname}
    \bibliography{main}

@String(AAAI = {AAAI})

@article{ha2018world,
  title={World models},
  author={Ha, David and Schmidhuber, J{\"u}rgen},
  journal={arXiv preprint arXiv:1803.10122},
  volume={2},
  number={3},
  pages={440},
  year={2018}
}

@article{hu2023gaia,
  title={Gaia-1: A generative world model for autonomous driving},
  author={Hu, Anthony and Russell, Lloyd and Yeo, Hudson and Murez, Zak and Fedoseev, George and Kendall, Alex and Shotton, Jamie and Corrado, Gianluca},
  journal={arXiv preprint arXiv:2309.17080},
  year={2023}
}

@inproceedings{wang2024driving,
  title={Driving into the future: Multiview visual forecasting and planning with world model for autonomous driving},
  author={Wang, Yuqi and He, Jiawei and Fan, Lue and Li, Hongxin and Chen, Yuntao and Zhang, Zhaoxiang},
  booktitle={Proceedings of the IEEE/CVF Conference on Computer Vision and Pattern Recognition},
  pages={14749--14759},
  year={2024}
}

@inproceedings{zheng2024occworld,
  title={Occworld: Learning a 3d occupancy world model for autonomous driving},
  author={Zheng, Wenzhao and Chen, Weiliang and Huang, Yuanhui and Zhang, Borui and Duan, Yueqi and Lu, Jiwen},
  booktitle={European conference on computer vision},
  pages={55--72},
  year={2024},
  organization={Springer}
}

@inproceedings{wang2024drivedreamer,
  title={Drivedreamer: Towards real-world-drive world models for autonomous driving},
  author={Wang, Xiaofeng and Zhu, Zheng and Huang, Guan and Chen, Xinze and Zhu, Jiagang and Lu, Jiwen},
  booktitle={European conference on computer vision},
  pages={55--72},
  year={2024},
  organization={Springer}
}

@inproceedings{zhang2024copilot4d,
  title={Copilot4d: Learning unsupervised world models for autonomous driving via discrete diffusion},
  author={Zhang, Lunjun and Xiong, Yuwen and Yang, Ze and Casas, Sergio and Hu, Rui and Urtasun, Raquel},
  booktitle={International Conference on Learning Representations},
  volume={2024},
  pages={7269--7299},
  year={2024}
}

@inproceedings{zyrianov2025lidardm,
  title={Lidardm: Generative lidar simulation in a generated world},
  author={Zyrianov, Vlas and Che, Henry and Liu, Zhijian and Wang, Shenlong},
  booktitle={2025 IEEE International Conference on Robotics and Automation (ICRA)},
  pages={6055--6062},
  year={2025},
  organization={IEEE}
}

@inproceedings{wu2026gem,
  title={GEM: Generating LiDAR World Model via Deformable Mamba},
  author={Wu, Yang and Liu, Zhaojiang and Meng, Qiang and Liu, Youquan and Weng, Renliang and Qian, Jianjun and Yang, Jian and Xie, Jin},
  booktitle={Proceedings of the IEEE/CVF Conference on Computer Vision and Pattern Recognition},
  pages={24227--24236},
  year={2026}
}

@article{gu2024dome,
  title={Dome: Taming diffusion model into high-fidelity controllable occupancy world model},
  author={Gu, Songen and Yin, Wei and Jin, Bu and Guo, Xiaoyang and Wang, Junming and Li, Haodong and Zhang, Qian and Long, Xiaoxiao},
  journal={arXiv preprint arXiv:2410.10429},
  year={2024}
}

@inproceedings{liao2025i2,
  title={I2-world: Intra-inter tokenization for efficient dynamic 4d scene forecasting},
  author={Liao, Zhimin and Wei, Ping and Zhang, Ruijie and Chen, Shuaijia and Wang, Haoxuan and Ren, Ziyang},
  booktitle={Proceedings of the IEEE/CVF International Conference on Computer Vision},
  pages={25810--25819},
  year={2025}
}

@article{tian2023occ3d,
  title={Occ3d: A large-scale 3d occupancy prediction benchmark for autonomous driving},
  author={Tian, Xiaoyu and Jiang, Tao and Yun, Longfei and Mao, Yucheng and Yang, Huitong and Wang, Yue and Wang, Yilun and Zhao, Hang},
  journal={Advances in Neural Information Processing Systems},
  volume={36},
  pages={64318--64330},
  year={2023}
}

@article{wang2026occsora,
  title={Occsora: 4d occupancy generation models as world simulators for autonomous driving},
  author={Wang, Lening and Zheng, Wenzhao and Ren, Yilong and Jiang, Han and Cui, Zhiyong and Yu, Haiyang and Lu, Jiwen},
  journal={IEEE Transactions on Image Processing},
  year={2026},
  publisher={IEEE}
}

@inproceedings{yang2025driving,
  title={Driving in the occupancy world: Vision-centric 4d occupancy forecasting and planning via world models for autonomous driving},
  author={Yang, Yu and Mei, Jianbiao and Ma, Yukai and Du, Siliang and Chen, Wenqing and Qian, Yijie and Feng, Yuxiang and Liu, Yong},
  booktitle={Proceedings of the AAAI Conference on Artificial Intelligence},
  volume={39},
  number={9},
  pages={9327--9335},
  year={2025}
}

@inproceedings{li2025uniscene,
  title={Uniscene: Unified occupancy-centric driving scene generation},
  author={Li, Bohan and Guo, Jiazhe and Liu, Hongsi and Zou, Yingshuang and Ding, Yikang and Chen, Xiwu and Zhu, Hu and Tan, Feiyang and Zhang, Chi and Wang, Tiancai and others},
  booktitle={Proceedings of the computer vision and pattern recognition conference},
  pages={11971--11981},
  year={2025}
}

@article{zhang2024d,
  title = {{D$^{2}$-World}: An Efficient World Model Through Decoupled Dynamic Flow},
  author={Zhang, Haiming and Yan, Xu and Xue, Ying and Guo, Zixuan and Cui, Shuguang and Li, Zhen and Liu, Bingbing},
  journal={arXiv preprint arXiv:2411.17027},
  year={2024}
}

@article{zhang2024efficient,
  title={An efficient occupancy world model via decoupled dynamic flow and image-assisted training},
  author={Zhang, Haiming and Xue, Ying and Yan, Xu and Zhang, Jiacheng and Qiu, Weichao and Bai, Dongfeng and Liu, Bingbing and Cui, Shuguang and Li, Zhen},
  journal={arXiv preprint arXiv:2412.13772},
  year={2024}
}

@article{shi2026come,
  title={Come: Adding scene-centric forecasting control to occupancy world model},
  author={Shi, Yining and Jiang, Kun and Meng, Qiang and Wang, Ke and Wang, Jiabao and Sun, Wenchao and Wen, Tuopu and others},
  journal={Advances in Neural Information Processing Systems},
  volume={38},
  pages={70--96},
  year={2025}
}

@article{li2024bevformer,
  title={Bevformer: learning bird’s-eye-view representation from lidar-camera via spatiotemporal transformers},
  author={Li, Zhiqi and Wang, Wenhai and Li, Hongyang and Xie, Enze and Sima, Chonghao and Lu, Tong and Yu, Qiao and Dai, Jifeng},
  journal={IEEE Transactions on Pattern Analysis and Machine Intelligence},
  volume={47},
  number={3},
  pages={2020--2036},
  year={2024},
  publisher={IEEE}
}

@article{huang2022bevdet4d,
  title={Bevdet4d: Exploit temporal cues in multi-camera 3d object detection},
  author={Huang, Junjie and Huang, Guan},
  journal={arXiv preprint arXiv:2203.17054},
  year={2022}
}

@inproceedings{yang2024visual,
  title={Visual point cloud forecasting enables scalable autonomous driving},
  author={Yang, Zetong and Chen, Li and Sun, Yanan and Li, Hongyang},
  booktitle={Proceedings of the IEEE/CVF Conference on Computer Vision and Pattern Recognition},
  pages={14673--14684},
  year={2024}
}

@inproceedings{caesar2020nuscenes,
  title={nuscenes: A multimodal dataset for autonomous driving},
  author={Caesar, Holger and Bankiti, Varun and Lang, Alex H and Vora, Sourabh and Liong, Venice Erin and Xu, Qiang and Krishnan, Anush and Pan, Yu and Baldan, Giancarlo and Beijbom, Oscar},
  booktitle={Proceedings of the IEEE/CVF conference on computer vision and pattern recognition},
  pages={11621--11631},
  year={2020}
}

@inproceedings{yang2023unisim,
  title={Unisim: A neural closed-loop sensor simulator},
  author={Yang, Ze and Chen, Yun and Wang, Jingkang and Manivasagam, Sivabalan and Ma, Wei-Chiu and Yang, Anqi Joyce and Urtasun, Raquel},
  booktitle={Proceedings of the IEEE/CVF Conference on Computer Vision and Pattern Recognition},
  pages={1389--1399},
  year={2023}
}

@inproceedings{bian2025dynamiccity,
  title={DynamicCity: Large-scale 4D occupancy generation from dynamic scenes},
  author={Bian, Hengwei and Kong, Lingdong and Xie, Haozhe and Pan, Liang and Qiao, Yu and Liu, Ziwei},
  booktitle={International Conference on Learning Representations},
  volume={2025},
  pages={40682--40716},
  year={2025}
}

@article{wei2024occllama,
  title={Occllama: An occupancy-language-action generative world model for autonomous driving},
  author={Wei, Julong and Yuan, Shanshuai and Li, Pengfei and Hu, Qingda and Gan, Zhongxue and Ding, Wenchao},
  journal={arXiv preprint arXiv:2409.03272},
  year={2024}
}

@inproceedings{xu2025occ,
  title={Occ-llm: Enhancing autonomous driving with occupancy-based large language models},
  author={Xu, Tianshuo and Lu, Hao and Yan, Xu and Cai, Yingjie and Liu, Bingbing and Chen, Yingcong},
  booktitle={2025 IEEE International Conference on Robotics and Automation (ICRA)},
  pages={8434--8441},
  year={2025},
  organization={IEEE}
}

@inproceedings{yan2025renderworld,
  title={Renderworld: World model with self-supervised 3d label},
  author={Yan, Ziyang and Dong, Wenzhen and Shao, Yihua and Lu, Yuhang and Liu, Haiyang and Liu, Jingwen and Wang, Haozhe and Wang, Zhe and Wang, Yan and Remondino, Fabio and others},
  booktitle={2025 IEEE International Conference on Robotics and Automation (ICRA)},
  pages={6063--6070},
  year={2025},
  organization={IEEE}
}

@inproceedings{liao2025stcocc,
  title={Stcocc: Sparse spatial-temporal cascade renovation for 3d occupancy and scene flow prediction},
  author={Liao, Zhimin and Wei, Ping and Chen, Shuaijia and Wang, Haoxuan and Ren, Ziyang},
  booktitle={Proceedings of the Computer Vision and Pattern Recognition Conference},
  pages={1516--1526},
  year={2025}
}

@article{loshchilov2017decoupled,
  title={Decoupled weight decay regularization},
  author={Loshchilov, Ilya and Hutter, Frank},
  journal={arXiv preprint arXiv:1711.05101},
  year={2017}
}
}
\end{document}